\documentclass{article}

\PassOptionsToPackage{numbers, compress}{natbib}
 \usepackage[preprint]{neurips_2026}

\usepackage[utf8]{inputenc} 
\usepackage[T1]{fontenc}    
\usepackage{hyperref}       
\usepackage{url}            
\usepackage{booktabs}       
\usepackage{amsfonts}       
\usepackage{nicefrac}       
\usepackage{microtype}      
\usepackage{xcolor}         
\usepackage[most]{tcolorbox}
\usepackage{graphicx}       
\usepackage[misc]{ifsym}
\usepackage{enumitem}
\usepackage{bm}
\usepackage{pifont, colortbl, booktabs}
\usepackage{makecell}
\usepackage{arydshln}
\usepackage{amsmath}
\usepackage{subcaption}
\usepackage{amssymb}
\usepackage{multirow}
\usepackage{listings}
\usepackage{wrapfig}
\usepackage{array}
\usepackage{titletoc}
\newcolumntype{C}[1]{>{\centering\arraybackslash}m{#1}}

\lstdefinestyle{prompt}{
  basicstyle=\footnotesize\ttfamily,
  breaklines=true,
  breakindent=1em,
  postbreak=\mbox{\ensuremath{\hookrightarrow}\ },
  columns=fullflexible,
  keepspaces=true,
  showstringspaces=false,
  xleftmargin=0.5em,
  aboveskip=4pt,
  belowskip=4pt,
  frame=none,
}

\definecolor{cYes}{HTML}{43A047}       
\definecolor{cPartial}{HTML}{1E88E5}   
\definecolor{cNo}{HTML}{D32F2F}        
\definecolor{rowOurs}{HTML}{E8F5E9}    
\definecolor{secHead}{HTML}{666666}    

\definecolor{rankgreenone}{HTML}{B7E4C7}
\definecolor{rankgreentwo}{HTML}{D8F3DC}
\definecolor{rankgreenthree}{HTML}{E8F5E9}

\newcommand{\rankone}[1]{\cellcolor{rankgreenone}\textbf{#1}}
\newcommand{\ranktwo}[1]{\cellcolor{rankgreentwo}\underline{#1}}
\newcommand{\rankthree}[1]{\cellcolor{rankgreenthree}#1}

\definecolor{takeawayblue}{HTML}{1F6FE5}
\definecolor{takeawaybg}{HTML}{EAF3FF}
\definecolor{takeawaytext}{HTML}{0B2545}

\newtcolorbox{takeawaybox}{
  colback=takeawaybg,
  colframe=takeawayblue,
  coltext=takeawaytext,
  boxrule=1.4pt,
  arc=5pt,
  left=10pt,
  right=10pt,
  top=8pt,
  bottom=8pt,
  width=\linewidth,
  enhanced,
}

\newcommand{\cmark}{\textcolor{cYes}{\ding{51}}}
\newcommand{\xmark}{\textcolor{cNo}{\ding{55}}}
\newcommand{\pmark}{\textcolor{cPartial}{\ding{72}}}

\definecolor{appleSoftRed}{HTML}{C96A73}
\definecolor{appleRose}{HTML}{D98778}
\definecolor{applePeach}{HTML}{E6A06D}
\definecolor{appleSoftGold}{HTML}{D9A84E}
\definecolor{appleLeafGreen}{HTML}{6F9E57}

\newcommand{\appleprojectlabel}{%
  \mbox{\normalfont\bfseries
  \textcolor{appleSoftRed}{P}%
  \textcolor{appleSoftRed}{r}%
  \textcolor{appleRose}{o}%
  \textcolor{appleRose}{j}%
  \textcolor{applePeach}{e}%
  \textcolor{applePeach}{c}%
  \textcolor{applePeach}{t}%
  \textcolor{appleSoftGold}{~P}%
  \textcolor{appleSoftGold}{a}%
  \textcolor{appleSoftGold}{g}%
  \textcolor{appleSoftGold}{e}%
  \textcolor{appleSoftGold}{:}}%
}

\newcommand{\appleprojecturl}[1]{%
  {\small
  \appleprojectlabel\quad
  \begingroup
  \hypersetup{pdfborder={0 0 0}}%
  \href{#1}{\textcolor{appleLeafGreen}{\nolinkurl{#1}}}%
  \endgroup}%
}

\title{\texorpdfstring{
  \vspace{-2pt}
  \noindent
  \makebox[\textwidth][c]{
    \smash{\raisebox{-0.55\height}{\includegraphics[height=2.3em]{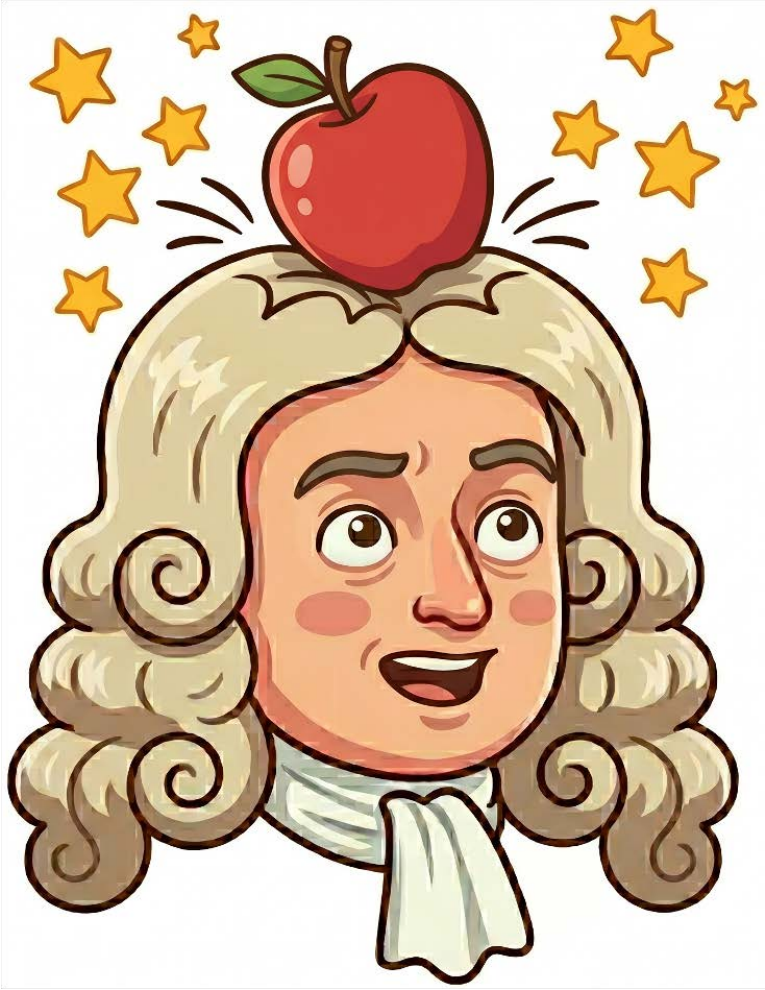}}}
    \hspace{0.2em}
    \parbox[c]{0.85\textwidth}{
        \centering 
        Apple-$\bm{\pi}$: Benchmarking Thinking with Video Towards Law-Grounded Physical Intelligence
        \vspace*{-15pt}
      }
  }
}{Apple-PI: Benchmarking Thinking with Video Towards Law-Grounded Physical Intelligence}
}

\author{
  Runmao Yao$^{*,1}$ \quad Kairui Hu$^{*,1}$ \quad Yukang Cao$^1$ \quad Ruisi Wang$^1$ \quad Shulin Tian$^1$ \\
  \quad \textbf{Ziang Cao$^1$} \quad \textbf{Weichen Fan$^1$} \quad \textbf{Ziqi Huang$^1$} \quad \textbf{Yuhao Dong$^1$} \quad  \textbf{Hao Li$^1$} \quad \textbf{Zhaoxi Chen$^1$}\\
  \textbf{Zhongang Cai$^1$} \quad \textbf{Lei Yang$^2$} \quad \textbf{Ziwei Liu}\textsuperscript{\hspace{0.1em}\Letter}$^{,1}$ \\
  $^{1}$S-Lab, Nanyang Technological University \quad $^{2}$The Chinese University of Hong Kong \\
  $^{*}$Equal contribution. \quad \textsuperscript{\Letter\hspace{0.1em}}Corresponding author. \\
  \appleprojecturl{https://21yrm.github.io/Apple-PI-homepage/}
}

\begin{document}

\maketitle

\vspace{-5pt}
\begin{figure*}[htbp]
\centering
\includegraphics[width=\textwidth]{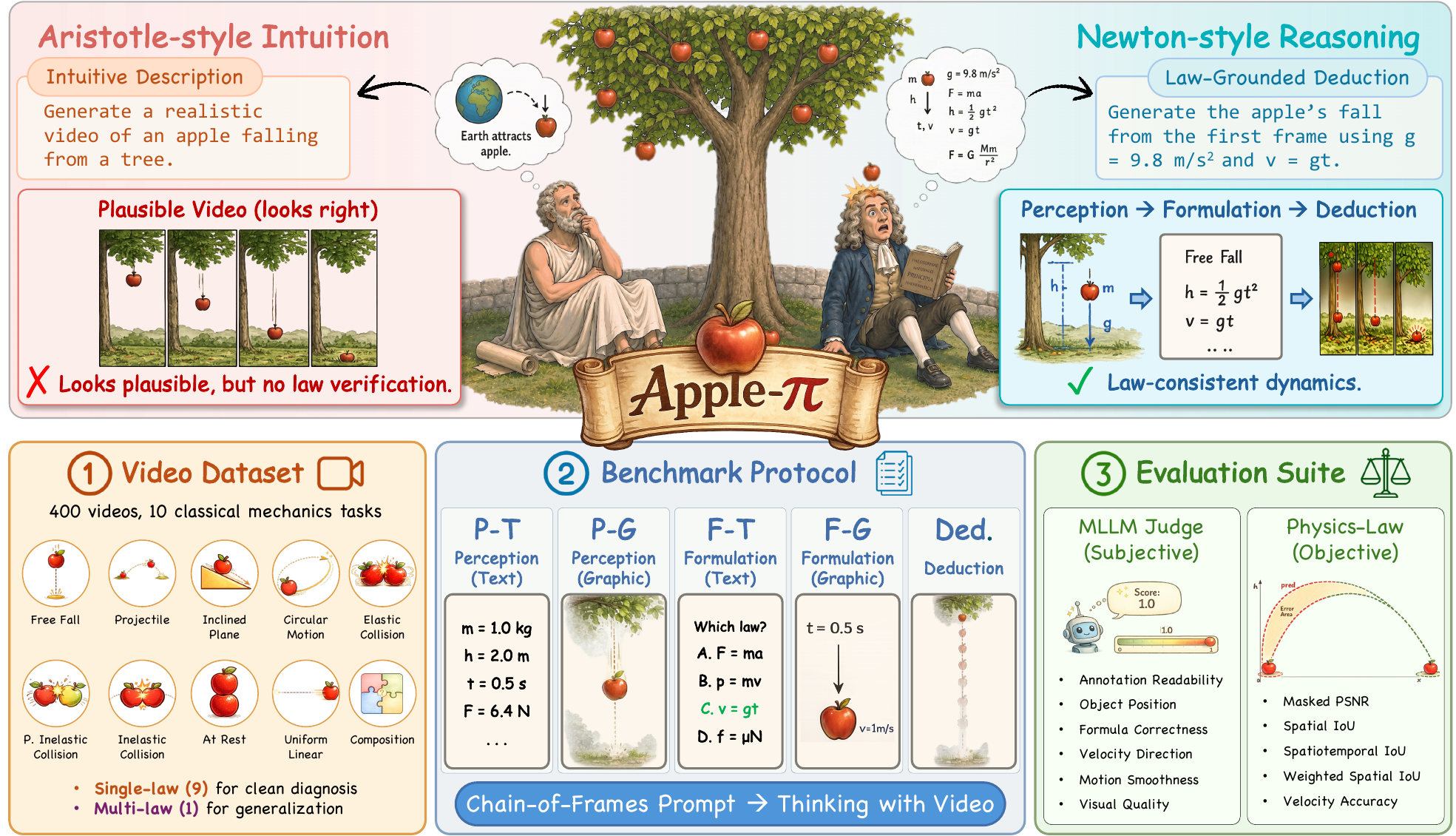}
\caption{\textbf{Apple-\(\bm\pi\) at a glance.}
Apple-\(\pi\) benchmarks law-grounded physical intelligence in video models through chain-of-frames traces over \textbf{Perception}, \textbf{Formulation}, and \textbf{Deduction}.
Built on the 400-video \textbf{Orchard} mechanics dataset, it combines a hybrid evaluation suite of MLLM-based and physics-law-grounded metrics to diagnose where reasoning fails.}
\label{fig:teaser}
\end{figure*}

\begin{abstract}
Modern video generation models are increasingly hailed as emerging world models with an internalized grasp of physical law. Yet existing benchmarks largely evaluate physical plausibility only at the output level, without verifying whether the model arrives there through a faithful, law-grounded reasoning process. We introduce \textbf{Apple-$\bm{\pi}$}, the first benchmark that anchors video-model evaluation explicitly in physical laws. Apple-$\pi$ comprises three components. \textbf{1) Orchard}: a dataset of 400 videos covering ten canonical tasks in classical mechanics. It separates single-law tasks for confounder-free diagnosis from multi-law tasks for probing generalization. \textbf{2) Benchmark Protocol}: a three-stage protocol based on scientific reasoning, including Perception, Formulation, and Deduction. It uses \emph{chain-of-frames} prompting on infographic-annotated first frames, treating the generated video as the model's visible reasoning trace. \textbf{3) Evaluation Suite}: a hybrid evaluation suite that combines MLLM-based subjective scoring with physics-law-grounded objective measures. This enables stage-resolved diagnosis of not only whether a model fails, but where it fails. Benchmarking 11 models shows that current video models remain far from reliable law-grounded world simulators, with the best video model scoring only 0.473. Our stage-, pillar-, and source-resolved analyses further expose a Perception-to-Formulation-to-Deduction bottleneck, weak multi-law state transfer, and a persistent Sim-to-Real gap. These findings position Apple-$\pi$ as a diagnostic foundation for guiding future video models toward world models with law-grounded physical intelligence.
\end{abstract}

\section{Introduction}

The history of physics traces a long trajectory from intuitive descriptions of natural phenomena to the formulation of the universal laws that govern them. Aristotle attributed a falling apple to its natural tendency to seek the earth. Newton, gazing at the same fall, abstracted a compact law, and from it deduced the motion of the moon, the tides, and the wandering planets. From a falling apple to the motion of the heavens, from the particular to the universal, from intuitive description to law-grounded deduction, this trajectory is the very signature of scientific reasoning.

Modern video models appear to be reenacting this trajectory at scale. Trained on massive collections of video, they implicitly absorb geometry, motion, and causality, accumulating an experience reminiscent of Aristotle's intuitive view of the world~\cite{wang2026bigvideoreasoningsuite}. Recent studies further suggest that video models can form a \emph{chain-of-frames} reasoning trace~\cite{ghazanfari2025chainofframesadvancingvideounderstanding}, and more broadly, may \emph{think with video}~\cite{tong2025thinkingvideovideogeneration}. These findings have fueled a growing conviction that video models are emerging as \emph{world models}~\cite{li2025worldmodelbenchjudgingvideogeneration}: that, through data alone, they have begun to internalize the governing laws of the physical world~\cite{garrido2025intuitivephysicsunderstandingemerges}.

Yet this bold claim remains difficult to verify with existing benchmarks. A growing body of work spans different laws~\cite{mak2026physicsmindsimrealmechanics, pmlr-v267-meng25c}, difficulty levels~\cite{bansal2025videophy2challengingactioncentricphysical, hu2025benchmarkingscientificunderstandingreasoning}, and evaluation protocols~\cite{gu2026phyworldbenchcomprehensiveevaluationphysical, Motamed_2026_WACV}, but all evaluate only what a model outputs, never how it got there. When a model succeeds, we cannot tell whether it truly invoked a physical law or merely produced something that looks right; when it fails, we cannot tell whether it misread the scene, misidentified the law, or stumbled in the deduction. In neither case can we answer the question that matters most: \emph{can video models reason about the physical world in a law-grounded manner, like Newton, rather than relying on intuition, like Aristotle?}

To make this question testable, \textbf{Apple-$\bm\pi$} turns Newton-style scientific reasoning into an auditable video-generation protocol: models must perceive the relevant physical quantities, formulate the governing law, and deduce law-consistent future dynamics. This design is instantiated through three tightly coupled components. \textbf{1) Video Dataset.} We curate \textbf{Orchard}, a collection of 400 physical videos drawn from real-world recordings, internet footage, and high-fidelity simulators. Orchard covers ten canonical tasks in classical mechanics, organized into single-law and multi-law levels to support both confounder-free diagnosis and compositional generalization. \textbf{2) Benchmark Protocol.} Every case in Orchard is evaluated along three tracks: \emph{Perception} tests whether a model can identify the physical scene, \emph{Formulation} tests whether it has internalized the governing law, and \emph{Deduction} tests whether it can produce law-consistent dynamics. Perception and Formulation are further split into \emph{text} and \emph{graphic} subtracks, yielding five subtracks in total. Across all subtracks, the model receives an infographic-style annotated first frame paired with a \emph{chain-of-frames} prompt, and responds with a video that visualizes its physical reasoning. \textbf{3) Evaluation Suite.} We design track-specific metrics that combine MLLM-based subjective scoring with physics-law-grounded objective measures, enabling fine-grained diagnosis of where in the reasoning chain a model breaks down.

Building on these components, we benchmark a broad range of state-of-the-art video generation and unified understanding-generation models. Our results show that current video models exhibit useful physical priors, but remain far from dependable law-grounded world models. They can often produce visually plausible motion from an annotated first frame, yet such plausibility does not necessarily indicate that the relevant physical quantities have been correctly grounded, the governing law has been internalized, or the resulting dynamics have been consistently followed over time. Unified understanding-generation models perform more strongly, suggesting that explicit understanding can play an important role in law-grounded video generation. Taken together, these findings position Apple-\(\pi\) as a diagnostic step toward the next generation of video models and world models that reason about the physical world not only visually, but also through explicit physical laws.

\section{Related Work}

\paragraph{Video World Models.} Learning world models from video has long been a goal in AI~\cite{ha2018recurrent, lecun2022path}, and Sora renewed this vision by framing large-scale video generation as a path toward world simulation~\cite{videoworldsimulators2024}. Recent advances in video generation~\cite{gao2025seedance10exploringboundaries, hong2022cogvideo, kong2025hunyuanvideosystematicframeworklarge, seedance2025seedance15pronative, seedance2026seedance20advancingvideo, wan2025wanopenadvancedlargescale, wiedemer2025videomodelszeroshotlearners, wu2025hunyuanvideo15technicalreport, yang2024cogvideox}, interactive world models~\cite{pmlr-v235-bruce24a, hyworld2026hyworld20multimodalworld, sun2025worldplaylongtermgeometricconsistency, robbyantteam2026advancingopensourceworldmodels}, and unified understanding-generation systems have made video an increasingly plausible substrate for physical simulation~\cite{cao2025physx3dphysicalgrounded3dasset, cao2025physxanythingsimulationreadyphysical3d, cao2026physxomniunifiedsimulationreadyphysical, Fang_2026_CVPR, klingteam2025klingomnitechnicalreport, 11094677, wei2026univideounifiedunderstandinggeneration, yao2026anchoreddreamzeroshot360degindoor}. Yet whether such models genuinely internalize governing physical laws, rather than replaying visual regularities, remains unresolved~\cite{pmlr-v267-kang25g}. Apple-$\pi$ targets this open question by evaluating video models as law-grounded physical reasoners, not merely as generators of visually plausible dynamics.

\begin{table*}[t]
\centering
\caption{\textbf{Comparison with representative physical intelligence benchmarks.} 
Apple-$\pi$ is the first to explicitly anchor evaluation in physical laws 
and decompose it into diagnosable reasoning stages. 
Source: R\,=\,Real, S\,=\,Simulated. Legend: \cmark\;full support\;\;\pmark\;partial\;\;\xmark\;not supported.}
\label{tab:bench_comparison}
\setlength{\tabcolsep}{5pt}
\resizebox{\textwidth}{!}{%
\begin{tabular}{@{\hskip 5pt} l c !{\color{black!15}\vrule width 0.6pt} c c c !{\color{black!15}\vrule width 0.6pt} c c c c @{\hskip 5pt}}
\toprule
\makecell[l]{\textbf{Benchmark}}
  & \makecell{\textbf{Eval Target}}
  & \makecell{\textbf{Data Source}}
  & \makecell{\textbf{\# Cases}}
  & \makecell{\textbf{Physics Scope}}
  & \makecell{\textbf{Law-}\\\textbf{Grounded}}
  & \makecell{\textbf{Multi-}\\\textbf{Stage}}
  & \makecell{\textbf{Chain-of-}\\\textbf{Frames}}
  & \makecell{\textbf{Objective}\\\textbf{Phys.\,Metric}} \\
\midrule
%
\rowcolor{black!6}
\multicolumn{9}{@{\hskip 5pt}l@{\hskip 5pt}}{\textit{\small\textcolor{secHead}{— Understanding Benchmarks —}}} \\
\noalign{\vspace{3pt}}
PhysBench~\cite{chow2025physbenchbenchmarkingenhancingvisionlanguage}
  & VLM & R+S & 10\,K & Multi-domain
  & \xmark & \xmark & \xmark & \xmark \\
QuantiPhy~\cite{puyin2025quantiphyquantitativebenchmarkevaluating}
  & VLM & R & 3.3\,K & Kinematics
  & \pmark & \xmark & \xmark & \cmark \\
PhyX~\cite{shen2025phyxdoesmodelwits}
  & VLM & R & 3\,K & Multi-domain
  & \xmark & \xmark & \xmark & \xmark \\
\midrule
%
\rowcolor{black!6}
\multicolumn{9}{@{\hskip 5pt}l@{\hskip 5pt}}{\textit{\small\textcolor{secHead}{— Generation Benchmarks —}}} \\
\noalign{\vspace{3pt}}
VideoPhy-2~\cite{bansal2025videophy2challengingactioncentricphysical}
  & T2V & R & 200 & Material interaction
  & \xmark & \xmark & \xmark & \xmark \\
PhyWorldBench~\cite{gu2026phyworldbenchcomprehensiveevaluationphysical}
  & T2V & R+S & 700 & Mechanics
  & \pmark & \xmark & \xmark & \xmark \\
VideoScience-Bench~\cite{hu2025benchmarkingscientificunderstandingreasoning}
  & T2V+I2V & R & 200 & Science
  & \pmark & \xmark & \xmark & \xmark \\
WorldModelBench~\cite{li2025worldmodelbenchjudgingvideogeneration}
  & T2V+I2V & R & 350 & Multi-domain
  & \xmark & \xmark & \xmark & \xmark \\
PhysicsMind~\cite{mak2026physicsmindsimrealmechanics}
  & VLM+I2V & R+S & 687 & Mechanics
  & \pmark & \xmark & \xmark & \cmark \\
PhyGenBench~\cite{pmlr-v267-meng25c}
  & T2V & — & 160 & Physical commonsense
  & \xmark & \xmark & \xmark & \xmark \\
Physics-IQ~\cite{Motamed_2026_WACV}
  & I2V & R & 396 & Multi-domain
  & \xmark & \xmark & \xmark & \cmark \\
\midrule
\rowcolor{rowOurs}
\textbf{Apple-$\bm{\pi}$\,(Ours)}
  & \textbf{I2V}
  & \textbf{R+S}
  & \textbf{400}
  & \textbf{Mechanics}
  & \cmark & \cmark & \cmark & \cmark \\
\bottomrule
\end{tabular}%
}
\end{table*}

\paragraph{Benchmarking Physical Intelligence.} A growing body of benchmarks targets physical intelligence in visual models; we summarize the most relevant efforts in Table~\ref{tab:bench_comparison}. On the \emph{understanding} side, benchmarks range from synthetic probes of intuitive physics~\cite{bear2022physionevaluatingphysicalprediction, bordes2025intphys2benchmarkingintuitive, riochet2020intphysframeworkbenchmarkvisual, yi2020clevrercollisioneventsvideo} to large-scale realistic QA suites covering diverse physics domains~\cite{chow2025physbenchbenchmarkingenhancingvisionlanguage, mak2026physicsmindsimrealmechanics, puyin2025quantiphyquantitativebenchmarkevaluating, shen2025phyxdoesmodelwits}. On the \emph{generation} side, VideoPhy~\cite{bansal2024videophyevaluatingphysicalcommonsense, bansal2025videophy2challengingactioncentricphysical} and Physics-IQ~\cite{Motamed_2026_WACV} assess physical plausibility through human ratings or pixel-level metrics, while others~\cite{gu2026phyworldbenchcomprehensiveevaluationphysical, hu2025benchmarkingscientificunderstandingreasoning, Huang_2024_CVPR, 11250949, le2025gravityvideogenerationposttraining, li2025worldmodelbenchjudgingvideogeneration, pmlr-v267-meng25c, wu2026omniworldbenchcomprehensiveinteractioncentricevaluation, zheng2025vbench20advancingvideogeneration,  zheng2026newtonbenchbenchmarkinggeneralizablescientific, zhou2026physinonevisualphysicslearning} further extend the coverage. Despite their breadth, existing benchmarks mostly evaluate whether outputs look physically plausible, without verifying the reasoning process behind them. Apple-$\pi$ fills this gap with explicit physical-law grounding, stage-resolved evaluation over \emph{Perception}, \emph{Formulation}, and \emph{Deduction}, and hybrid MLLM-based subjective and physics-law-grounded objective metrics.

\paragraph{Thinking with Video.} Chain-of-thought prompting has evolved from \emph{thinking with text}~\cite{NEURIPS2022_9d560961} to \emph{thinking with image}~\cite{chen2026unitunifiedmultimodalchainofthought, NEURIPS2024_fb820110, openai2025thinkingwithimages, zhang2026thinkstrokespixelsprocessdriven}, and most recently to \emph{thinking with video}~\cite{chen2025tivibenchbenchmarkingthinkinvideoreasoning, guo2025videomodelsreadyzeroshot, li2026frameprocessawareevaluationgenerative, luo2025vreasonbenchunifiedreasoningbenchmark, tong2026coft2ivideomodelspure, wang2026demystifingvideoreasoning}. Wiedemer et al.~\cite{wiedemer2025videomodelszeroshotlearners} formalize this latest step as \emph{chain-of-frames}, drawing an explicit analogy between frame-by-frame video generation and step-by-step textual reasoning. Subsequent work has expanded this direction across benchmarking, generation, and test-time scaling~\cite{huang2025vchainchainofvisualthoughtreasoningvideo, li2026thinkingframesvisualcontext, Liu_2025_ICCV, tong2025thinkingvideovideogeneration}. Apple-$\pi$ leverages this paradigm for evaluation: our \emph{chain-of-frames} prompts elicit visible, frame-level reasoning traces, making the model's physical thought process auditable across all tracks.
\section{Apple-\texorpdfstring{$\bm{\pi}$}{π}}

Apple-$\pi$ tests whether video models can move beyond intuitive physical plausibility toward law-grounded physical intelligence. It consists of three coupled components: \textbf{Orchard}, a dataset of law-derived, physically specified scenarios (\S~\ref{sec:orchard}); a \textbf{benchmark protocol} probing reasoning through \emph{Perception}, \emph{Formulation}, and \emph{Deduction} (\S~\ref{sec:protocol}); and an \textbf{evaluation suite} combining MLLM-based subjective and physics-law-grounded objective metrics to diagnose where models fail (\S~\ref{sec:eval_suite}).

\subsection{Orchard: Video Dataset}
\label{sec:orchard}

Existing physical video datasets often contain realistic but entangled phenomena, making model failures difficult to attribute. Orchard takes a law-first approach: each case is organized around explicit classical-mechanics laws, physically specified conditions, and law-predicted motion. This design yields analyzable and reproducible videos for stage-resolved evaluation, while retaining controlled real-world visual diversity across sources. Orchard contains 400 cases drawn from complementary data sources and organized by a two-level task taxonomy, as detailed below.

\begin{figure*}[h]
\centering
\includegraphics[width=\textwidth]{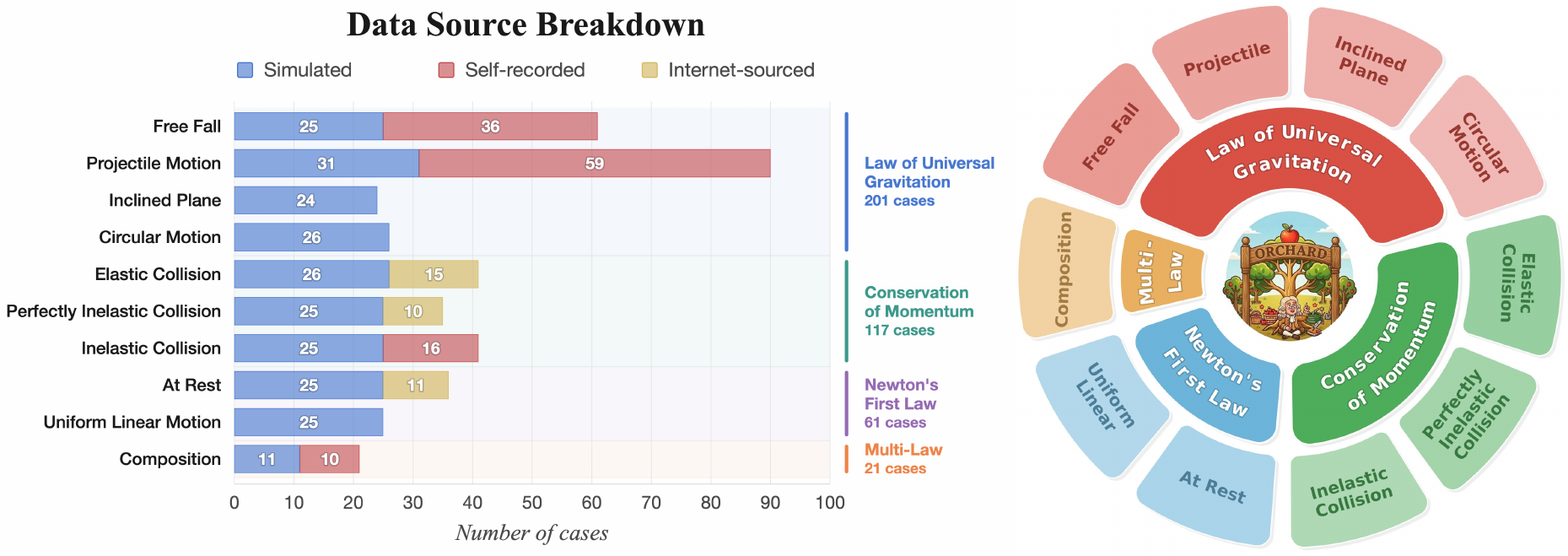}
\caption{
\textbf{Overview of Orchard.} Left: Per-task data-source breakdown across simulated, self-recorded, and Internet-sourced videos. Right: Two-level task taxonomy organizing 400 cases into three single-law pillars and a multi-law composition branch.
}
\label{fig:orchard_overview}
\end{figure*}

\paragraph{Data sources.} Orchard draws from three complementary origins, with the per-task source breakdown shown in the left panel of Figure~\ref{fig:orchard_overview}. \emph{Simulated videos} (243 cases), generated with NVIDIA Isaac Sim~\cite{NVIDIA_Isaac_Sim}, give us exact physical parameters and pixel-accurate trajectories, anchoring the objective measures of the evaluation suite. \emph{Self-recorded real-world videos} (121 cases), captured under controlled laboratory conditions, approximate isolation of the governing law while introducing the optical and material cues absent from simulation. \emph{Internet-sourced real-world videos} (36 cases), curated from physics-education channels on YouTube, broaden visual diversity to settings the first two cannot cover. Across all three sources, we control object identity as a nuisance 
variable by standardizing the object vocabulary to four primitive 
solids: sphere, cube, cylinder, and cone. Their simple geometry yields 
well-defined centers, masks, contact surfaces, and size parameters, 
which makes physical states consistently annotatable across simulation, 
self-recorded videos, and Internet-sourced videos. This design also reduces the chance that model performance is driven by object-specific semantic priors, keeping the benchmark focused on law-grounded motion reasoning.

\paragraph{Task taxonomy.} As shown in the right panel of Figure~\ref{fig:orchard_overview}, Orchard follows a two-level task taxonomy. At the top level, we divide cases into 
a \emph{single-law} branch for controlled diagnosis and a \emph{multi-law} 
branch for compositional generalization. The single-law branch is further 
organized into three pillars, each defined by the dominant physical principle 
it probes. The first pillar, \emph{law of universal gravitation}, includes free 
fall ($h=\tfrac{1}{2}gt^2$), projectile motion 
($\vec{r}(t)=\vec{r}_0+\vec{v}_0t+\tfrac{1}{2}\vec{g}t^2$), inclined-plane 
motion ($a=g\sin\theta$), and circular motion on a gravity-driven track 
($mg=mv^2/r$). The second pillar, \emph{conservation of momentum}, 
includes perfectly elastic collisions, perfectly inelastic collisions, and inelastic collisions, parameterized by the coefficient of restitution $e=1$, $e=0$, 
and $0<e<1$, respectively. The third pillar, \emph{Newton's first law}, 
includes objects at rest and objects in uniform linear motion under 
$\sum \vec{F}=0$. The multi-law branch composes tasks across these pillars, 
such as an inclined plane feeding into projectile motion or circular motion 
followed by a collision, to test whether models can chain laws that they may 
handle individually.

\begin{figure*}[t]
\centering
\includegraphics[width=\textwidth]{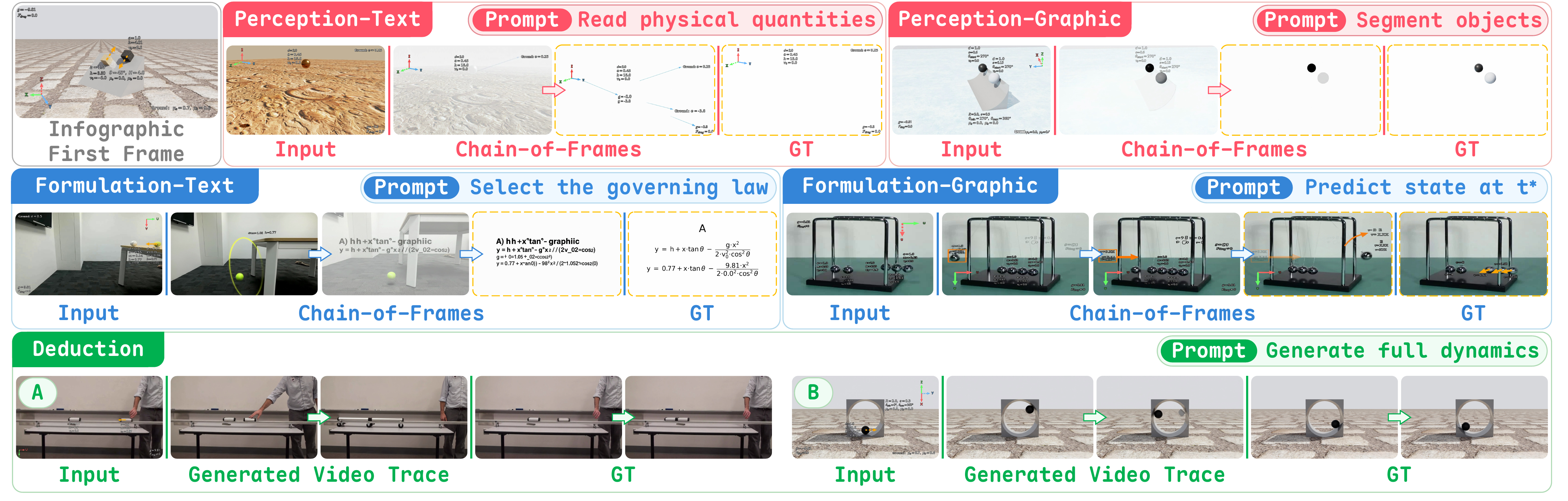}
\caption{
\textbf{Apple-$\bm{\pi}$ benchmark protocol.}
An infographic-annotated first frame and chain-of-frames prompt elicit five reasoning subtracks: Perception-Text for reading quantities, Perception-Graphic for segmenting objects, Formulation-Text for selecting laws, Formulation-Graphic for predicting target states, and Deduction for generating full dynamics.
}
\label{fig:protocol}
\end{figure*}

\subsection{Benchmark Protocol}
\label{sec:protocol}

Apple-$\pi$ evaluates each Orchard case through three stages of scientific reasoning: Perception, Formulation, and Deduction 
(Figure~\ref{fig:protocol}). Perception and Formulation are each 
split into text and graphic subtracks, while Deduction is evaluated 
as a single video generation subtrack, yielding five subtracks in 
total. All five subtracks share a common input format: an annotated 
first frame paired with a \emph{chain-of-frames} text prompt. They 
also share a common output format: a generated video that serves as 
the model's visible reasoning trace. We first detail this shared 
input and output paradigm (\S~\ref{sec:io_format}), then specify what 
each subtrack asks of the model (\S~\ref{sec:eval_tracks}).

\subsubsection{Input and Output Format}
\label{sec:io_format}

\paragraph{Input format: infographic-style first frame.}
Before a model can reason about a physical scene, it must first bind 
each physical quantity to the correct visual referent. Text prompts make 
this binding unnecessarily indirect. In a scene with multiple objects, 
the model must determine which object a mass or initial velocity 
describes; in an inclined plane scene, it must determine which surface an 
angle or friction coefficient describes. This reference-resolution burden is not the target of our benchmark. We therefore provide 
the parameters of each Orchard case as \textbf{infographic-style overlays} on the 
first frame, placed next to their visual referents. Environmental 
constants, such as gravitational acceleration, 
appear as global labels, while object properties, such as mass, appear adjacent to the corresponding objects. See Appendix~\ref{app:design_principles} for a detailed rationale.

\paragraph{Output format: chain-of-frames video.}
Apple-$\pi$ uses video as the common output format across all subtracks. We prompt models with a \emph{chain-of-frames} instruction that evolves the infographic-annotated first frame toward the required answer, either as a full sequence or as a final-frame answer artifact. See Appendix~\ref{app:protocol_prompts} for prompt templates. Depending on the subtrack, the response takes one of three forms: \textbf{1)} the full generated sequence, used when the entire trajectory is evaluated; \textbf{2)} a final-frame artifact with fade-to-white~\cite{chen2026frameplacevideocontent}, where the video fades to a clean white background containing only the requested discrete answer, such as a label or numeric value; and \textbf{3)} a final-frame artifact without fade-to-white, used when the answer must remain grounded in the original scene, such as predicted object positions overlaid on the environment.

\subsubsection{Stage and Subtrack Specification}
\label{sec:eval_tracks}

\paragraph{Perception.}
Perception tests the foundational stage of reasoning: identifying 
the physical entities present in the scene, before any law can be 
invoked. We decompose this stage into two complementary sub-skills, 
distinguished by the type of entity recognized. 
\textbf{1) Perception-Text} targets physical \emph{quantities}, 
analogous to OCR: the model reproduces the numeric annotations 
from the input frame, delivered as a final-frame artifact 
with fade-to-white that preserves each annotation at its original 
spatial position. \textbf{2) Perception-Graphic} targets physical 
\emph{objects}, analogous to instance segmentation: the model 
localizes the experiment-relevant objects, also delivered as a 
final-frame artifact with fade-to-white, with target 
objects preserved unchanged while the surrounding environment 
fades out.

\paragraph{Formulation.}
Formulation tests the abstraction stage of reasoning: whether 
the model has internalized the law that ties the perceived 
quantities together. Following the two natural modes in which a 
law can be expressed, we split this stage into two complementary 
sub-skills. \textbf{1) Formulation-Text} tests the law as a 
\emph{symbolic equation}: the model selects from a four-option 
multiple-choice question containing the \emph{correct} law 
and three distractors designed to expose specific failure modes:
a \emph{confusing} real law that shares symbols with the annotations,
an \emph{unrelated} real law with no relevant symbol overlap,
and a \emph{fabricated} formula that does not exist in physics. The response is delivered as a final-frame 
artifact with fade-to-white, bearing three lines on the white 
background: the chosen option label, the symbolic formula, and 
the formula with each annotated symbol substituted by its numeric 
value. \textbf{2) Formulation-Graphic} tests the law as a 
\emph{predicted state}: given a target instant $t^\star$ 
specified in the prompt (e.g., ``at $t^\star=2\,$s''), the model 
predicts the configuration of the scene at $t^\star$. The 
response is delivered as a final-frame artifact without 
fade-to-white, where each object appears at its $t^\star$ 
position in the original scene, overlaid with a velocity arrow 
and a speed label.

\paragraph{Deduction.}
Deduction tests the final stage of reasoning: producing 
law-consistent dynamics over time. Given the annotated first 
frame, the model generates a complete video that simulates how 
the scene evolves under the governing law. The response is the full generated sequence, evaluated frame by frame against the law-predicted ground-truth trajectory.

\subsection{Evaluation Suite}
\label{sec:eval_suite}

Apple-$\pi$ scores every subtrack on a $[0,1]$ scale using two complementary axes: 
\emph{MLLM-based subjective scoring} for response validity and format compliance, and 
\emph{physics-law-grounded objective measures} for law consistency. 
The former applies to all subtracks, while the latter applies only when object regions or trajectories can be matched against physics-derived ground truth. 
Concrete metric definitions are deferred to Appendix~\ref{app:standardization} and Appendix~\ref{app:metrics}.

\paragraph{MLLM-based subjective scoring.}
Visual qualities such as content, layout, and style cannot be 
verified by physics alone; we delegate them to an MLLM judge 
equipped with a track-specific rubric. The rubric's fine-grained 
criteria are organized into semantically coherent groups (e.g., 
\emph{formula correctness} and \emph{symbol substitution} for 
Formulation-Text), and each subtrack's final score is a weighted 
average across groups, with weights reflecting each group's 
importance to that subtrack.

\paragraph{Physics-law-grounded objective measures.}
Law-grounded physical correctness is the central property 
Apple-$\pi$ aims to evaluate, and some subtracks naturally admit 
direct evaluation against physics-derived ground truth. 
Perception-Graphic and Formulation-Graphic both deliver predicted 
object regions, scored by a segmentation-IoU measure against 
ground-truth masks. Deduction, whose response is a full trajectory, is evaluated with a richer battery: pixel-level fidelity, spatiotemporal mask overlap, and 3D velocity 
error, all measured against the law-predicted ground-truth 
dynamics.
\section{Experiments}
\label{sec:experiments}

\subsection{Experimental Setup}

We evaluate 11 representative models on the full Apple-$\pi$ benchmark, including 5 video generation models (Wan2.2~\cite{wan2025wanopenadvancedlargescale}, HunyuanVideo-1.5~\cite{wu2025hunyuanvideo15technicalreport}, VBVR-Wan2.2~\cite{wang2026bigvideoreasoningsuite}, Seedance~2.0~\cite{seedance2026seedance20advancingvideo}, and Veo~3.1~\cite{wong2026veo31ingredients}) and 6 unified understanding-generation models (BAGEL~\cite{deng2025emergingpropertiesunifiedmultimodal}, OmniGen2~\cite{wu2026omnigen2instructionalignedmultimodalgeneration}, SenseNova-U1-8B-MoT~\cite{opensensenova2026sensenovau1}, SenseNova-U1-8B-MoT-Think~\cite{opensensenova2026sensenovau1}, GPT Image 2~\cite{openai2026chatgptimages20}, and Nano Banana 2~\cite{google2026nanobanana2}). Each model is tested on 400 cases, five subtracks, and three independent rollouts, yielding $400 \times 5 \times 3 = 6000$ evaluated responses per model. We include unified models as an architectural contrast to video models, since their explicit coupling of visual understanding and generation allows us to probe whether understanding-centric design improves law-grounded physical intelligence. For the four non-Deduction subtracks that are evaluated as final-frame artifacts, unified models directly generate the required final-frame outputs. For Deduction, where the target output is a full video sequence, we ask unified models to generate a sparse set of keyframes at the evaluation timestamps and score them using the corresponding frame-level criteria. We use Gemini~3~Flash as the MLLM judge. Further implementation details are provided in Appendix~\ref{app:standardization}, \ref{app:metrics}, and \ref{app:experiment_release}.

\begin{table}[t]
\centering
\caption{
\textbf{Main results.}
Scores are reported on a $[0,1]$ scale, with higher being better, and are grouped by reasoning track,
physical-law pillar, and data source.
P-T/P-G denote Perception-Text/Graphic; F-T/F-G denote Formulation-Text/Graphic.
Grav., Mom., N1, and Multi denote law of universal gravitation, conservation of momentum, Newton's first law, and multi-law cases.
}
\label{tab:main_results}
\setlength{\tabcolsep}{4pt}
\setlength{\cmidrulekern}{0.35em}
\resizebox{\linewidth}{!}{%
\begin{tabular}{@{\hskip 4pt}l c c c c c c c c c c C{1.24cm} C{1.24cm} @{\hskip 4pt}}
\toprule
\multirow{2}{*}{{\textbf{Model}}}
& \multirow{2}{*}{\makecell[c]{\textbf{Avg.}}}
& \multicolumn{5}{c}{\textbf{Track-wise Score}}
& \multicolumn{4}{c}{\textbf{Pillar-wise Score}}
& \multicolumn{2}{c}{\textbf{Source-wise Score}} \\
\cmidrule(lr){3-7} \cmidrule(lr){8-11} \cmidrule(lr){12-13}
&
& \textbf{P-T}
& \textbf{P-G}
& \textbf{F-T}
& \textbf{F-G}
& \multicolumn{1}{c!{\color{black!15}\vrule width 0.6pt}}{\textbf{Ded.}}
& \textbf{Grav.}
& \textbf{Mom.}
& \textbf{N1}
& \multicolumn{1}{c!{\color{black!15}\vrule width 0.6pt}}{\textbf{Multi}}
& \textbf{Sim.}
& \textbf{Real} \\
\midrule

\rowcolor{black!6}
\multicolumn{13}{@{\hskip 4pt}l@{\hskip 4pt}}{
\textit{\small\textcolor{black!65}{— Video Models —}}
} \\
\noalign{\vspace{3pt}}

Wan2.2~\cite{wan2025wanopenadvancedlargescale}
  & 0.267
  & 0.645 & 0.257 & 0.009 & 0.275
  & \multicolumn{1}{c!{\color{black!15}\vrule width 0.6pt}}{0.149}
  & 0.245 & 0.274 & 0.344
  & \multicolumn{1}{c!{\color{black!15}\vrule width 0.6pt}}{0.200}
  & 0.310 & 0.224 \\

HunyuanVideo-1.5~\cite{wu2025hunyuanvideo15technicalreport}
  & 0.177
  & 0.230 & 0.292 & 0.027 & 0.180
  & \multicolumn{1}{c!{\color{black!15}\vrule width 0.6pt}}{0.155}
  & 0.128 & 0.205 & 0.272
  & \multicolumn{1}{c!{\color{black!15}\vrule width 0.6pt}}{0.183}
  & 0.208 & 0.146 \\

VBVR-Wan2.2~\cite{wang2026bigvideoreasoningsuite}
& 0.373
& \ranktwo{0.923} & \rankthree{0.502} & 0.001 & 0.239
& \multicolumn{1}{c!{\color{black!15}\vrule width 0.6pt}}{0.201}
& 0.347 & 0.375 & 0.459
& \multicolumn{1}{c!{\color{black!15}\vrule width 0.6pt}}{0.387}
& 0.394 & 0.352 \\

Seedance 2.0~\cite{seedance2026seedance20advancingvideo}
& \rankthree{0.473}
& 0.597 & 0.490 & \rankthree{0.478} & \rankthree{0.485}
& \multicolumn{1}{c!{\color{black!15}\vrule width 0.6pt}}{\rankthree{0.315}}
& \rankthree{0.450} & \rankthree{0.510} & \rankthree{0.495}
& \multicolumn{1}{c!{\color{black!15}\vrule width 0.6pt}}{\rankthree{0.389}}
& \rankthree{0.487} & \rankthree{0.459} \\

Veo 3.1~\cite{wong2026veo31ingredients}
& 0.313
& 0.408 & 0.357 & 0.325 & 0.316
& \multicolumn{1}{c!{\color{black!15}\vrule width 0.6pt}}{0.160}
& 0.288 & 0.314 & 0.420
& \multicolumn{1}{c!{\color{black!15}\vrule width 0.6pt}}{0.235}
& 0.356 & 0.270 \\

\midrule

\rowcolor{black!6}
\multicolumn{13}{@{\hskip 4pt}l@{\hskip 4pt}}{
\textit{\small\textcolor{black!65}{— Unified Models —}}
} \\
\noalign{\vspace{3pt}}

BAGEL~\cite{deng2025emergingpropertiesunifiedmultimodal}
  & 0.218
  & 0.218 & 0.383 & 0.091 & 0.186
  & \multicolumn{1}{c!{\color{black!15}\vrule width 0.6pt}}{0.213}
  & 0.192 & 0.242 & 0.275
  & \multicolumn{1}{c!{\color{black!15}\vrule width 0.6pt}}{0.162}
  & 0.246 & 0.191 \\

OmniGen2~\cite{wu2026omnigen2instructionalignedmultimodalgeneration}
 & 0.214
 & 0.168 & 0.442 & 0.061 & 0.213
 & \multicolumn{1}{c!{\color{black!15}\vrule width 0.6pt}}{0.184}
 & 0.168 & 0.232 & 0.313
 & \multicolumn{1}{c!{\color{black!15}\vrule width 0.6pt}}{0.208}
 & 0.242 & 0.185 \\

SenseNova-U1-8B-MoT~\cite{opensensenova2026sensenovau1}
& 0.362
& 0.515 & 0.492 & 0.157 & 0.339
& \multicolumn{1}{c!{\color{black!15}\vrule width 0.6pt}}{0.304}
& 0.326 & 0.389 & 0.454
& \multicolumn{1}{c!{\color{black!15}\vrule width 0.6pt}}{0.267}
& 0.393 & 0.330 \\

SenseNova-U1-8B-MoT-Think~\cite{opensensenova2026sensenovau1}
& 0.359
& 0.505 & 0.485 & 0.154 & 0.345
& \multicolumn{1}{c!{\color{black!15}\vrule width 0.6pt}}{0.307}
& 0.322 & 0.390 & 0.449
& \multicolumn{1}{c!{\color{black!15}\vrule width 0.6pt}}{0.268}
& 0.390 & 0.328 \\

GPT Image 2~\cite{openai2026chatgptimages20}
& \rankone{0.704}
& \rankthree{0.921} & \rankone{0.719} & \ranktwo{0.824} & \rankone{0.651}
& \multicolumn{1}{c!{\color{black!15}\vrule width 0.6pt}}{\rankone{0.406}}
& \rankone{0.695} & \ranktwo{0.711} & \rankone{0.742}
& \multicolumn{1}{c!{\color{black!15}\vrule width 0.6pt}}{\rankone{0.618}}
& \ranktwo{0.740} & \rankone{0.668} \\

Nano Banana 2~\cite{google2026nanobanana2}
& \ranktwo{0.699}
& \rankone{0.934} & \ranktwo{0.667} & \rankone{0.841} & \ranktwo{0.650}
& \multicolumn{1}{c!{\color{black!15}\vrule width 0.6pt}}{\ranktwo{0.405}}
& \ranktwo{0.684} & \rankone{0.723} & \ranktwo{0.733}
& \multicolumn{1}{c!{\color{black!15}\vrule width 0.6pt}}{\ranktwo{0.567}}
& \rankone{0.743} & \ranktwo{0.656} \\

\bottomrule
\end{tabular}%
}
\vspace{-10pt}
\end{table}

\subsection{Main Results}

Table~\ref{tab:main_results} reports the overall performance of all evaluated models on Apple-\(\pi\). The benchmark remains challenging for current video generation models. The best video model, Seedance 2.0, reaches an average score of 0.473, while other video models remain substantially lower. In contrast, GPT Image 2 and Nano Banana 2 achieve the highest overall scores, 0.704 and 0.699, respectively. This gap suggests that strong video synthesis alone does not yet imply reliable law-grounded physical intelligence: models must not only produce plausible motion, but also bind explicit quantities, identify governing laws, and generate law-consistent dynamics.

\vspace{-5pt}
\paragraph{Large-scale video training and reasoning supervision.} The video-model group reveals two partial sources of law-grounded physical intelligence. First, proprietary models such as Seedance 2.0 and Veo 3.1 generally outperform base open-source generators such as Wan2.2 and HunyuanVideo-1.5, suggesting that large-scale, high-quality video training can distill useful physical priors. However, their absolute scores remain far from saturated, showing that such priors are insufficient under explicit quantities, laws, and target states. Second, VBVR-Wan2.2, a Wan2.2 variant fine-tuned on the video-reasoning VBVR-Dataset~\cite{wang2026bigvideoreasoningsuite}, performs competitively among video models, especially on Perception-Text and Perception-Graphic. This suggests that targeted reasoning supervision improves interface skills such as annotation reading and object grounding, but its low Formulation-Text and Deduction scores show that these skills do not yet transfer reliably to law formulation or dynamic deduction. Overall, video-model results suggest that both large-scale video pretraining and reasoning-oriented supervision help, but neither alone solves Apple-\(\pi\).

\vspace{-5pt}
\paragraph{Explicit understanding before generation.} The unified-model group provides a second perspective on the source of law-grounded physical intelligence. GPT Image 2 and Nano Banana 2 substantially outperform all video models across most tracks, especially Perception and Formulation, suggesting that explicit understanding and controllable visual generation are beneficial for reading annotations, grounding objects, selecting laws, and producing structured final-frame answers. However, open-source unified models remain much weaker, indicating that the advantage does not come from the unified interface alone. It likely also depends on stronger multimodal foundations, broader training data, and mature post-training for instruction following, text rendering, layout control, and controllable generation. Moreover, even the strongest unified models score only around 0.40 on Deduction, indicating that law-consistent temporal dynamics remains the central bottleneck. These results suggest that future video world models may need not only stronger temporal generation, but also explicit mechanisms for physical understanding.

\begin{takeawaybox}
\textbf{Takeaway 1}\quad
Large-scale video training provides useful physical priors, but reliable law-grounded physical intelligence appears to require explicit understanding modules, richer physical-reasoning data, and reasoning-oriented post-training.
\end{takeawaybox}

\begin{figure*}[t]
\centering
\includegraphics[width=\textwidth]{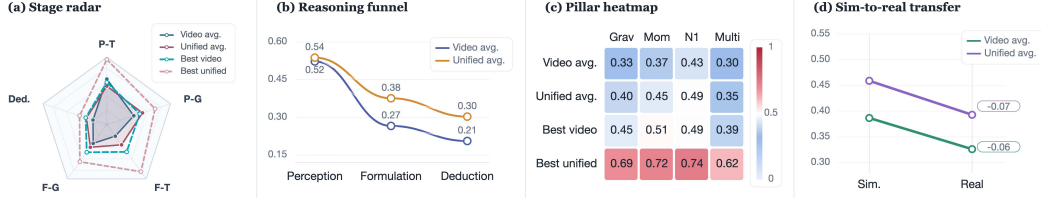}
\caption{
\textbf{Stage-, pillar-, and source-resolved analysis.}
(a,b) Stage-wise results show a decline from Perception to Formulation to Deduction.
(c) Pillar-wise results compare single- and multi-law cases.
(d) Source-wise results show the Sim-to-Real gap.
}
\vspace{-10pt}
\label{fig:exp}
\end{figure*}

\subsection{Stage-Resolved Diagnosis}

\paragraph{Progressive stage-wise bottleneck.}
Apple-$\pi$'s stage-wise design reveals where video models fail along the physical reasoning chain. Focusing on the video-model group, Table~\ref{tab:main_results} and Figure~\ref{fig:exp}(a) show a clear difficulty order: Perception is easiest, Formulation is harder, and Deduction is hardest. This hierarchy follows the protocol: Perception mainly requires recovering physical entities from the annotated first frame, Formulation requires abstracting or instantiating the governing law, and Deduction requires rolling that law forward into full temporal dynamics. Thus, video models do not fail uniformly; they often capture local visual or textual cues, but struggle once those cues must be converted into law-conditioned temporal evolution.

\paragraph{Subtrack-level failure modes.}
The subtracks expose finer-grained bottlenecks within each stage. In Perception, Perception-Text is generally easier than Perception-Graphic, suggesting that reading visible quantities is easier than grounding them to object regions. In Formulation, Formulation-Text and Formulation-Graphic reveal a gap between symbolic law selection and grounded state prediction: a model may choose a plausible equation without correctly placing objects or velocities at the target time. In Deduction, the challenge becomes fully temporal, requiring models to maintain object identity, velocity, acceleration, and contact events across frames; visually smooth motion can still violate the law-derived trajectory or collision response.

\paragraph{Reasoning funnel.}
Figure~\ref{fig:exp}(b) summarizes this pattern as a reasoning funnel: average scores drop from Perception to Formulation and further to Deduction. Earlier success is therefore necessary but not sufficient. Good Perception does not guarantee correct Formulation, since copying annotations can remain superficial without identifying the governing law. Likewise, good Formulation does not guarantee correct Deduction, since selecting a law or predicting a single target state is easier than executing that law consistently over time. Unified models provide a useful reference, showing that explicit understanding modules help across stages but do not remove the deduction bottleneck.

\begin{takeawaybox}
\textbf{Takeaway 2}\quad
Video models show a progressive reasoning bottleneck: they can often perceive
physical cues and partially formulate laws, but these intermediate successes do not reliably
transfer to law-consistent dynamics over time.
\end{takeawaybox}

\subsection{Pillar-wise and Source-wise Analysis}

\paragraph{Generalization across laws.}
The pillar-wise results compare three single-law pillars, Grav., Mom., and N1, with the multi-law branch. As shown in Table~\ref{tab:main_results} and Figure~\ref{fig:exp}(c), Multi is generally harder than the single-law pillars. This gap reflects the compositional nature of multi-law cases: the state produced by one law, such as position, velocity, direction, or contact, must become the initial condition for the next. The lower Multi scores suggest that current video models can capture isolated law-specific motion priors, but remain weak at carrying physical states across law transitions, where state transfer and temporal consistency become critical.

\paragraph{Generalization across sources.}
The source-wise results reveal a complementary limitation. As shown in Figure~\ref{fig:exp}(d), both video and unified models perform worse on real-world cases than on simulated cases, although unified models maintain higher absolute scores. Since the governing laws remain unchanged across sources, this Sim-to-Real drop mainly reflects failures in grounding and tracking under realistic visual conditions. Thus, stronger understanding modules help, but do not remove the need for robust generalization across visual domains.

\begin{takeawaybox}
\textbf{Takeaway 3}\quad
Video models show limited generalization across both laws and sources: they
struggle to carry physical states across law transitions in multi-law cases and to apply the
same laws under real-world visual variation.
\end{takeawaybox}

\subsection{Qualitative Failure Analysis}

\begin{wrapfigure}{r}{0.45\textwidth}
  \centering
  \includegraphics[width=0.43\textwidth]{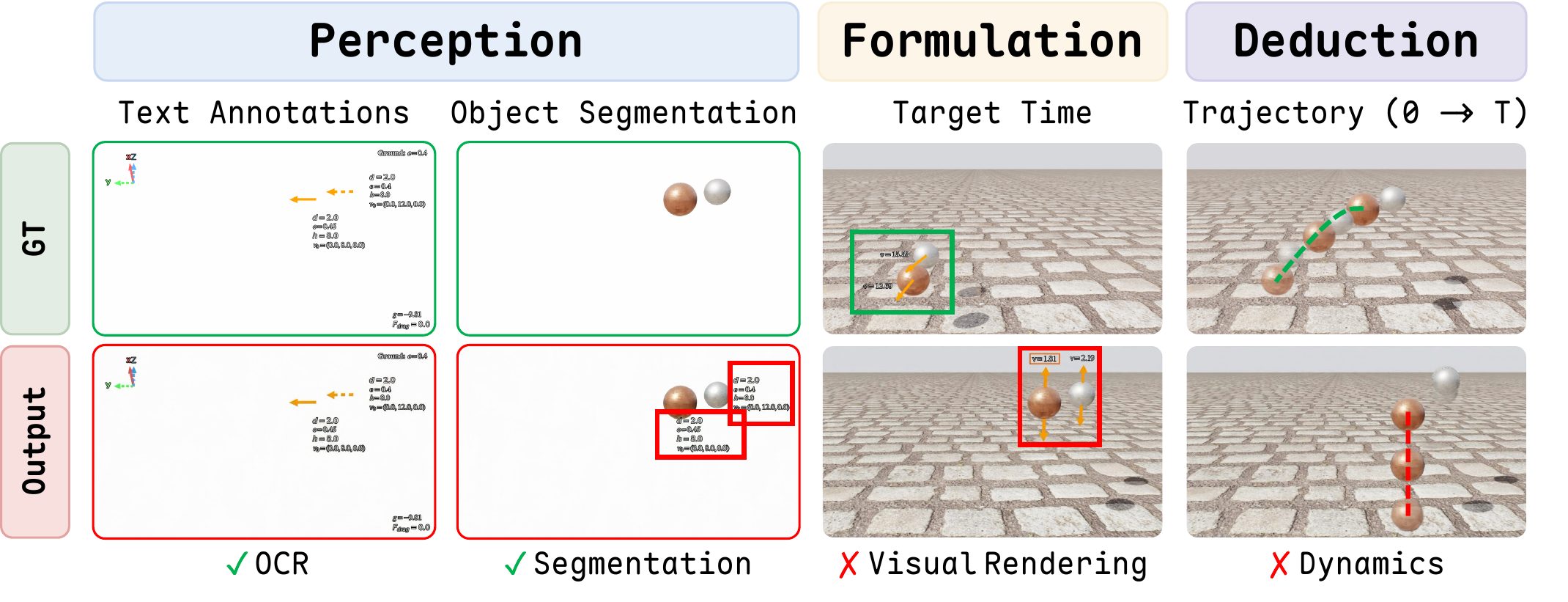}
\caption{
\textbf{Qualitative failure analysis.}
The model preserves annotations and objects, but misbinds the initial-velocity cue, yielding a wrong target state and trajectory.
}
  \label{fig:failure}
\end{wrapfigure}

To complement the quantitative results, Figure~\ref{fig:failure} visualizes representative failure cases. A common failure is annotation semantics rather than perception: the model may preserve the text and track the object, yet fail to bind an annotation to its physical meaning. For example, an initial-velocity arrow can be visually recognized but assigned the wrong direction as a physical initial condition, causing both the Formulation-Graphic target state and the Deduction trajectory to evolve incorrectly. This shows that copying annotations does not imply understanding the underlying physical variables.

A second class of failures occurs before physics reasoning begins. Some models fail to follow the requested output format, such as blurring annotation text, failing to isolate target objects, producing unreadable formulas, or omitting velocity arrows and labels. These errors reflect limitations in OCR, segmentation, rendering, and instruction following.

\subsection{Protocol Ablations}

We ablate two protocol choices in Apple-$\pi$: how physical quantities are specified and how detailed the task prompt should be. Table~\ref{tab:protocol_ablation} reports paired deltas against the default setting.

\begin{wraptable}{r}{0.52\textwidth}
\vspace{-4mm}
\centering
\caption{\textbf{Protocol ablations.} Values are paired score deltas from the default setting; positive means better, and ``\textemdash'' marks inapplicable subtracks.}
\label{tab:protocol_ablation}
\setlength{\tabcolsep}{2.5pt}
\renewcommand{\arraystretch}{0.88}
\scriptsize
\resizebox{\linewidth}{!}{%
\begin{tabular}{l c c c c c c}
\toprule
\multirow{2}{*}{\textbf{Model}}
& \multirow{2}{*}{\textbf{Avg. $\Delta$}}
& \multicolumn{5}{c}{\textbf{Track-wise $\Delta$}} \\
\cmidrule(lr){3-7}
&
& \textbf{P-T}
& \textbf{P-G}
& \textbf{F-T}
& \textbf{F-G}
& \textbf{Ded.} \\
\midrule

\rowcolor{black!6}
\multicolumn{7}{l}{
\textit{\textcolor{black!65}{— Text-Parameter —}}
} \\

Veo 3.1~\cite{wong2026veo31ingredients}
& +0.008
& \textemdash
& +0.020
& +0.038
& -0.024
& +0.000 \\

Nano Banana 2~\cite{google2026nanobanana2}
& +0.020
& \textemdash
& +0.044
& -0.006
& +0.011
& +0.030 \\

\midrule

\rowcolor{black!6}
\multicolumn{7}{l}{
\textit{\textcolor{black!65}{— Concise-Prompt —}}
} \\

Veo 3.1~\cite{wong2026veo31ingredients}
& -0.018
& +0.015
& -0.068
& +0.051
& -0.096
& +0.006 \\

Nano Banana 2~\cite{google2026nanobanana2}
& -0.037
& -0.016
& -0.085
& +0.051
& -0.089
& -0.045 \\

\bottomrule
\end{tabular}%
}
\vspace{-4mm}
\end{wraptable}

\vspace{-5pt}
\paragraph{Text-Parameter.} Providing physical quantities in a structured text prompt is nearly neutral: the average change is small for both models, and all available track-level deltas stay within $\pm 0.05$. This indicates that Apple-$\pi$ is not primarily bottlenecked by reading infographic annotations. We therefore keep infographic annotations not as a performance shortcut, but as a compact, spatially grounded, and auditable way to bind each physical quantity to its visual referent.

\vspace{-5pt}
\paragraph{Concise-Prompt.} A concise track-level prompt with less detailed output specifications slightly lowers the average score. The drop is concentrated on graphic-output tracks, especially P-G and F-G, suggesting that the full prompt helps enforce visual constraints. Meanwhile, F-T improves by +0.051 for both models, showing that concise instructions are sufficient for formula selection. Overall, the full prompt mainly standardizes the visual answer format rather than injecting physics solutions.
\section{Conclusion}

We introduced Apple-\(\pi\), a law-grounded benchmark for evaluating whether video generation models can move beyond visual physical plausibility toward auditable physical intelligence. Built on Orchard, Apple-\(\pi\) anchors each case in explicit classical-mechanics laws and decomposes evaluation into Perception, Formulation, and Deduction. Experiments show that current video models acquire useful physical priors from large-scale video training, but remain far from reliable law-grounded simulators, especially when quantities must be formulated into laws, executed over time, composed across law transitions, and transferred to real-world domains. Stronger unified models suggest that future video world models will require not only better temporal generation, but also explicit understanding modules, richer physical-reasoning data, and reasoning-oriented post-training.

\begin{ack}
This study is supported by the Ministry of Education, Singapore, under its MOE AcRF Tier 2 (MOE-T2EP20223-0002). This research is also supported by cash and inkind funding from NTU S-Lab and industry partner(s).
\end{ack}

{
    \small
    \bibliographystyle{myplainnat}
    \bibliography{references}
}

\clearpage
\appendix
\startcontents[appendix]
\section*{Appendix Contents}
\printcontents[appendix]{}{1}{}

\providecommand{\applepi}{Apple-$\pi$}
\providecommand{\orchard}{Orchard}

\section{Orchard Design Principles and Data Card}
\label{app:orchard}

\subsection{Design Principles}
\label{app:design_principles}

\paragraph{Law-first rather than video-first.}
\orchard \ is curated from a law-first perspective. Instead of starting from arbitrary videos and
retrospectively assigning physical labels, each case begins with an explicit governing principle in
classical mechanics, a specified initial condition, and a law-predicted future trajectory. This design
makes \applepi \ diagnostic: when a model fails, we can localize the failure to Perception,
Formulation, or Deduction rather than treating the generated video as an undifferentiated endpoint.

\paragraph{Single-law diagnosis and multi-law generalization.}
The dataset is divided into a single-law branch and a multi-law branch. Single-law cases isolate one
dominant physical principle, enabling clean diagnosis with minimal confounding. Multi-law cases
compose two or more principles, such as an inclined-plane stage followed by projectile motion or a
curved-track stage followed by collision. These cases test whether a model can transfer state variables
from one law to the next: position, velocity, contact status, object identity, and direction of motion.

\paragraph{Controlled object vocabulary.}
Across simulated, self-recorded, and Internet-sourced cases, we standardize the dynamic objects to
four primitive solids: sphere, cube, cylinder, and cone. This does not make the benchmark purely
synthetic or visually trivial. Rather, it controls object-specific semantic priors and yields consistent
centers, masks, contact surfaces, and size parameters, which are necessary for reproducible
law-grounded evaluation.

\paragraph{Spatially grounded physical quantities.}
Physical quantities are attached to visual referents. A mass belongs to an object; a friction coefficient
belongs to a contact surface; an angle belongs to a ramp; a velocity vector belongs to a moving body
and direction. We therefore provide an infographic-style annotated first frame. The annotations
specify initial conditions but not the answer: models must still identify the governing law and deduce
the future state or trajectory. This interface reduces irrelevant reference-resolution burden while
preserving the intended physical reasoning challenge.

\subsection{Dataset Composition and Native Frame Rates}
\label{app:dataset_composition}

\orchard \ contains 400 cases spanning ten canonical mechanics tasks. The dataset combines three
sources with different ground-truth construction procedures and native frame rates, summarized in
Table~\ref{tab:source_fps}. Importantly, simulator and real-world cases are not treated as identical:
simulator cases provide programmatic ground truth from the physics engine, while real-world cases
provide measurement-backed and human-reviewed annotations.

\begin{table}[b]
\centering
\small
\caption{Data sources and native ground-truth frame rates. Frame rate is stored per case and used by
the evaluation pipeline; \applepi \ does not assume a single global GT fps.}
\label{tab:source_fps}
\begin{tabular}{p{0.20\linewidth}p{0.12\linewidth}p{0.18\linewidth}p{0.43\linewidth}}
\toprule
Source & Cases & Native GT fps & Ground-truth source \\
\midrule
Simulated & 243 & 24 fps & Isaac Sim/PhysX state, render passes, masks, depth, object pose,
velocity, contact events, and analytically verified labels. \\
Self-recorded real & 121 & 30 fps & Controlled laboratory recordings with instrumented measurements,
human-refined masks, and physics-derived state annotations. \\
Internet-sourced real & 36 & Native fps of each video & Curated physics-demonstration clips with per-video fps stored in
metadata; only clips with auditable motion windows and sufficient annotation information are retained. \\
\bottomrule
\end{tabular}
\end{table}

\subsection{Task Taxonomy}
\label{app:taxonomy}

Table~\ref{tab:task_taxonomy} lists the task families. The formula column gives the representative
law used to construct or verify the case; individual instances may include task-specific parameters
such as non-zero initial velocity, friction, restitution, or non-Earth gravitational acceleration.

\begin{table}[t]
\centering
\small
\caption{Task taxonomy in \orchard. The first nine tasks are single-law diagnostic tasks; the last row
contains multi-law compositions.}
\label{tab:task_taxonomy}
\begin{tabular}{p{0.20\linewidth}p{0.22\linewidth}p{0.25\linewidth}p{0.25\linewidth}}
\toprule
Task & Pillar & Representative law & Evaluated physical state \\
\midrule
Free fall & Universal gravitation & $v(t)=v_0+gt$, $h(t)=h_0+v_0t+\frac{1}{2}gt^2$ &
Vertical position and velocity. \\
Projectile motion & Universal gravitation & $\mathbf r(t)=\mathbf r_0+\mathbf v_0t+\frac{1}{2}\mathbf g t^2$ &
2D/3D trajectory and velocity direction. \\
Inclined plane & Universal gravitation & $a=g(\sin\theta-\mu\cos\theta)$, with $\mu=0$ for frictionless cases &
Ramp-aligned acceleration and object position. \\
Circular motion & Universal gravitation & $mg \approx mv^2/r$ or energy-derived speed along the track &
Track following, tangent velocity, and contact. \\
Perfectly elastic collision & Conservation of momentum & Momentum conservation with $e=1$ &
Post-collision velocity and identity preservation. \\
Perfectly inelastic collision & Conservation of momentum & Momentum conservation with $e=0$ &
Merged motion and shared velocity. \\
Inelastic collision & Conservation of momentum & Momentum conservation with $0<e<1$ &
Partially dissipative collision response. \\
At rest & Newton's first law & $\sum \mathbf F=0,\ \mathbf v=0$ &
Static object and absence of drift/deformation. \\
Uniform linear motion & Newton's first law & $\sum \mathbf F=0,\ \mathbf v(t)=\mathbf v_0$ &
Constant-speed straight-line motion. \\
Composition & Multi-law & Sequential composition of the above laws &
State transfer across law transitions. \\
\bottomrule
\end{tabular}
\end{table}

\subsection{Released Metadata}
\label{app:metadata}

Each case is released as a self-contained directory. Table~\ref{tab:metadata_schema} summarizes the
canonical files. Source-specific fields are optional, but the evaluation-critical fields are always present:
native GT fps, physical duration, timestamp list, object masks when available, formula metadata, and
target-time state.

\begin{table}[t]
\centering
\small
\caption{Canonical case-level metadata and artifacts.}
\label{tab:metadata_schema}
\begin{tabular}{p{0.30\linewidth}p{0.62\linewidth}}
\toprule
Artifact & Description \\
\midrule
\texttt{first\_frame.png} & Unannotated first frame, when available. \\
\texttt{annotated\_first\_frame.png} & Infographic-style input frame used for all five subtracks. \\
\texttt{gt/video.mp4} or \texttt{gt/rgb/} & Ground-truth video or RGB frame directory. \\
\texttt{gt/masks/} & Per-frame object masks for objective IoU and masked metrics. \\
\texttt{case.json} & Source type, task type, object list, native GT fps, dimensions, camera setting, and source notes. \\
\texttt{physics\_duration.txt} & Requested physical duration $T$ in seconds. \\
\texttt{timestamps.json} & Canonical evaluation timestamps $\mathcal T_c$. \\
\texttt{formula\_info.json} & Four-way formula candidates, correct option, symbolic formula, and substituted values. \\
\texttt{velocity.json} & Target time $t^\star$, object centers, velocity vectors, speeds, and velocity-label metadata. \\
\texttt{measurement.json} & Real-world measurement records, when applicable. \\
\bottomrule
\end{tabular}
\end{table}

\section{Ground Truth, Annotation, and Quality Assurance}
\label{app:annotation}

\subsection{Simulator Ground Truth}
\label{app:sim_gt}

Simulator cases are generated in Isaac Sim with deterministic case configurations. Each configuration
specifies dynamic objects, primitive shape, material, geometry, initial pose, initial linear/angular
velocity, environment parameters, ground properties, camera pose, and rendering options. The
simulator subset is programmatic-first: RGB frames, masks, depth, object poses, velocities, and
contact events are read from the engine state or render passes rather than manually estimated.

Two consistency rules are used during case authoring. First, mass is derived from material density and
analytic primitive volume, $m=\rho V$, so changing a shape or material updates the physical mass
consistently. Second, visual material and physics material share a single material key, preventing a
case from having a visually metallic object with unrelated density, friction, or restitution parameters.
When a task admits a closed-form solution, such as ideal collision velocities or projectile states, the
analytic value is used as the canonical answer key and the engine state is retained for diagnostics.

\subsection{Real-World Ground Truth}
\label{app:real_gt}

Real-world cases are measurement-first rather than engine-state-first. Self-recorded clips are captured
at 30 fps under controlled laboratory conditions. Internet-sourced clips retain their native fps, which is
stored in the case metadata. For both real-world sources, physical parameters are obtained through
instrumented measurements, visible calibration cues, public object specifications when appropriate,
and human review. The goal is not to make real videos identical to simulated videos, but to keep the
governing law auditable under realistic appearance, lighting, texture, and camera conditions.

\begin{table}[t]
\centering
\small
\caption{Real-world physical-parameter measurement. The velocity procedure uses the recorded fps:
30 fps for self-recorded clips and the native fps for Internet-sourced clips.}
\label{tab:measurement}
\begin{tabular}{p{0.32\linewidth}p{0.36\linewidth}p{0.20\linewidth}}
\toprule
Quantity & Instrument or procedure & Precision / convention \\
\midrule
Object mass & Digital balance & $\pm 0.01$ g \\
Object size & Digital caliper & $\pm 0.05$ mm \\
Drop / launch height & Steel tape plus frame-interpolation check & $\pm 1$ mm \\
Initial velocity & Frame-by-frame velocimetry at recorded fps & $\pm 0.05$ m/s \\
Coefficient of restitution & Two-drop ratio test on the surface & $\pm 0.02$ \\
Friction coefficient & Inclined-plane tilt-threshold test & $\pm 0.05$ \\
Gravitational acceleration & Case metadata; default Earth value for ordinary real-world clips & Stored per case \\
\bottomrule
\end{tabular}
\end{table}

\subsection{Mask and Tracking Annotation}
\label{app:mask_annotation}

Object masks are generated with a semi-automatic pipeline. Initial masks are produced by a segmentation
model, propagated through the clip, and then inspected by human annotators. If the automatic masks
show drift, over-segmentation, under-segmentation, or identity switches, annotators correct the mask
at the failing frame and re-run propagation. For multi-object cases, object identity is checked across
the full evaluation window, with special attention to contact and collision frames.

For simulator cases, engine-rendered instance masks are canonical. For real-world cases, the
human-refined masks are canonical. These masks support both graphic-subtrack IoU and Deduction
trajectory metrics.

\subsection{Formula Answer-Key Construction}
\label{app:formula_keys}

For each case, \texttt{formula\_info.json} contains four candidate formulas:
\begin{enumerate}
\item the correct governing formula for the case;
\item a confusing real formula that shares symbols with the annotations but does not govern the case;
\item an unrelated real formula from another mechanics family;
\item a fabricated formula that is syntactically plausible but not a valid physical law.
\end{enumerate}
This construction distinguishes law selection from superficial symbol matching. A model that chooses
a formula merely because it contains the same variables should fail on the confusing distractor.

\subsection{Three-Pass Review Protocol}
\label{app:qa}

Every case undergoes a 1+2 review protocol: one primary annotation pass followed by two independent
secondary review passes. Reviewers check the initial frame, annotation legibility, physical duration,
target time, mask identity, velocity labels, formula candidates, and source metadata. Disagreements
are resolved in the following order: analytic physical backstop when available; senior arbitration for
visual or measurement ambiguities; domain-expert escalation for formula-choice ambiguity.

Table~\ref{tab:catch_rate} reports issues caught by a complete simulator-subset review pass. These
corrections indicate that the review protocol is not merely procedural but actively improves label
quality.

\begin{table}[t]
\centering
\small
\caption{Examples of issues caught and corrected during simulator QA.}
\label{tab:catch_rate}
\begin{tabular}{p{0.58\linewidth}p{0.20\linewidth}}
\toprule
Fix category & Cases caught \\
\midrule
Physics-duration adjustment & 23 \\
Velocity-label theoretical-value override & 10 \\
Configuration re-recording due to camera, material, or position issue & 30 \\
\bottomrule
\end{tabular}
\end{table}

\subsection{Inter-Annotator Agreement}
\label{app:iaa}

We conduct a retroactive inter-annotator agreement study across simulator and real-world categories.
Two production annotators independently re-annotate held-out cases under blind conditions. Categorical
fields are measured with Cohen's $\kappa$; continuous fields are measured with ICC(2,1) after
quantization to the protocol grid or instrument precision. Results are shown in
Tables~\ref{tab:iaa_cat} and~\ref{tab:iaa_cont}.

\begin{table}[t]
\centering
\small
\caption{Inter-annotator agreement on categorical fields.}
\label{tab:iaa_cat}
\begin{tabular}{p{0.55\linewidth}p{0.20\linewidth}}
\toprule
Field & Cohen's $\kappa$ \\
\midrule
Object naming convention compliance & 0.97 \\
Task-type classification & 1.00 \\
Formula correct-answer choice & 0.96 \\
Camera-framing acceptability & 0.91 \\
Velocity-label legibility & 0.89 \\
\bottomrule
\end{tabular}
\end{table}

\begin{table}[t]
\centering
\small
\caption{Inter-annotator agreement on continuous fields.}
\label{tab:iaa_cont}
\begin{tabular}{p{0.55\linewidth}p{0.20\linewidth}}
\toprule
Field & ICC(2,1) \\
\midrule
Physics duration & 0.94 \\
Target time & 0.96 \\
Real-world object mass & 0.99 \\
Real-world drop / launch height & 0.98 \\
Real-world coefficient of restitution & 0.88 \\
Real-world initial velocity & 0.95 \\
\bottomrule
\end{tabular}
\end{table}

\section{Benchmark Protocol and Prompt Templates}
\label{app:protocol_prompts}

\subsection{Five Subtracks}
\label{app:subtracks}

All subtracks receive the same annotated first frame and a subtrack-specific chain-of-frames prompt.
The expected output differs by subtrack, as summarized in Table~\ref{tab:subtracks}.

\begin{table}[t]
\centering
\small
\caption{Input-output specification for the five \applepi \ subtracks.}
\label{tab:subtracks}
\begin{tabular}{p{0.16\linewidth}p{0.24\linewidth}p{0.31\linewidth}p{0.20\linewidth}}
\toprule
Subtrack & Question & Expected output & Main evaluation target \\
\midrule
P-T & Read physical quantities & Final white-background annotation artifact & Text/OCR, annotation placement, readability \\
P-G & Ground physical objects & Final white-background object-only artifact & Object localization and mask IoU \\
F-T & Select governing law & Final white-background three-line formula answer & Option, formula, substitution \\
F-G & Predict state at $t^\star$ & Scene at target time with velocity arrows and labels & Target-state mask IoU, arrow direction, speed label \\
Deduction & Generate full dynamics & Full trajectory video or timestamped keyframes & Law-consistent temporal evolution \\
\bottomrule
\end{tabular}
\end{table}

\subsection{Prompt Families}
\label{app:prompt_families}

We use two prompt families because evaluated models expose different interfaces. Unified
understanding-generation models produce still images for the four non-Deduction subtracks and
timestamped keyframes for Deduction. Video-generation models produce an MP4 for every subtrack;
for the four non-Deduction subtracks, the last frame of the generated video is used as the answer
artifact.

Runtime placeholders include \texttt{\{formula\_choices\}}, \texttt{\{target\_time\}},
\texttt{\{time\_point\}}, and \texttt{\{physics\_duration\}}. The prompts below are reproduced
from the released code's machine-readable templates, \texttt{UM\_PROMPT\_TEMPLATES} and
\texttt{VM\_PROMPT\_TEMPLATES}; source em dashes are rendered as \texttt{--} for
\texttt{listings} compatibility under pdfLaTeX.

\subsubsection{Unified Understanding-Generation Models}
\label{app:prompts_unified}

Unified models receive a single still-image directive per non-Deduction subtrack. For the
Deduction subtrack, they receive the per-keyframe directive below once per requested timestamp.

\paragraph{Perception-Text.}\leavevmode
\begin{lstlisting}[style=prompt]
Given this physics infographic, generate a new image that shows ONLY the human-drawn annotations on a pure white background. Keep all text labels, coordinate axes, and velocity arrows exactly as they appear -- same position, content, color, font, and size. Text must be crisp and readable. Remove everything else: the background, ground, sky, and all physical objects. Do not add any extra elements not in the original.
\end{lstlisting}

\paragraph{Perception-Graphic.}\leavevmode
\begin{lstlisting}[style=prompt]
Show only the subject physical objects from this image on a pure white background. Remove all annotations (text labels, coordinate axes, velocity arrows), the ground surface, sky, and supporting structures (ramps, tracks, rails, platforms). Keep only the objects that have labeled physical properties. Do not transform, resize, or replace any object -- a cube must stay a cube, a sphere must stay a sphere. Each object must retain its exact position, size, shape, color, and texture.
\end{lstlisting}

\paragraph{Formulation-Text.}\leavevmode
The runtime placeholder \texttt{\{formula\_choices\}} is filled with the case's four candidate
formulas read from \texttt{formula\_info.json}.
\begin{lstlisting}[style=prompt]
This physics infographic shows a scene with annotated physical parameters. Below are 4 candidate formulas. Exactly ONE is the correct governing equation for the motion shown:
{formula_choices}

Generate an image with a pure white background containing THREE lines of large, clearly readable, centered text:
  Line 1: The correct option letter (A/B/C/D).
  Line 2: The correct formula exactly as written above.
  Line 3: The formula with annotated variables replaced by their numeric values from the image, keeping unannotated symbols unchanged.

When substituting: if multiple objects exist, pick exactly one consistent set of values.
\end{lstlisting}

\paragraph{Formulation-Graphic.}\leavevmode
The runtime placeholder \texttt{\{target\_time\}} is the case-specific target instant $t^\star$.
\begin{lstlisting}[style=prompt]
Predict the physical state at t = {target_time} s. Generate an image showing the scene at that instant, where the original annotations (text labels, coordinate axes, arrows) have been cleanly removed while the background, ground, sky, and all scene lighting remain completely unchanged from the input. Show each object at its correct position at t = {target_time} s -- each object must retain its exact texture, material, size, and shape. For each moving object, overlay a borderless orange arrow starting from the object's center pointing in its instantaneous velocity direction, with a label "v = X.XX" (speed value, no unit). If an object is at rest, show "v = 0" with no arrow. Keep the same viewpoint as the input image -- no zoom, no rotation, no crop.
\end{lstlisting}

\paragraph{Deduction.}\leavevmode
\label{app:prompts_unified_ded}
The runtime placeholder \texttt{\{time\_point\}} is filled with the requested timestamp for each
keyframe.
\begin{lstlisting}[style=prompt]
This image is an annotated physics scene. All annotations follow SI units: g is gravitational acceleration, e is restitution (0-1), v0 is initial velocity (arrows show direction), mu is friction coefficient. Generate what the scene looks like at t = {time_point} seconds after motion begins, following these physical parameters accurately. Remove all annotations (text, axes, arrows). Objects are rigid bodies -- same shape, size, color, texture throughout. Static camera, fixed background and ground.
\end{lstlisting}

\subsubsection{Video-Generation Models}
\label{app:prompts_videogen}

Video-generation models receive a chain-of-frames directive that elicits one video. For the four
non-Deduction subtracks, the prompt asks the video to settle on the answer artifact and the last frame
is used for scoring; for Deduction, the full generated video is scored.

\paragraph{Perception-Text.}\leavevmode
\begin{lstlisting}[style=prompt]
Generate a video where the background, ground, sky, and all physical objects gradually fade to pure white, while all human-drawn annotations (text labels, coordinate axes, velocity arrows) remain completely unchanged. Static camera, no zoom, no rotation. The final frame should show only the original annotations on white -- crisp, sharp, and readable. Do not add any extra annotations or elements not in the original.
\end{lstlisting}

\paragraph{Perception-Graphic.}\leavevmode
\begin{lstlisting}[style=prompt]
Generate a video where the background, ground, sky, and all annotations (text labels, coordinate axes, velocity arrows) gradually fade to pure white, while the subject physical objects remain still and unchanged. Static camera, no zoom, no rotation. The final frame should show only the objects with labeled physical properties on white -- remove supporting structures (ramps, tracks, rails, platforms). Do not transform, resize, or replace any object. Each object must retain its exact position, size, shape, color, and texture.
\end{lstlisting}

\paragraph{Formulation-Text.}\leavevmode
As in the unified-model variant, \texttt{\{formula\_choices\}} is filled from
\texttt{formula\_info.json}.
\begin{lstlisting}[style=prompt]
This physics infographic shows annotated physical parameters describing a specific type of motion. Below are 4 candidate formulas. Exactly ONE is correct:
{formula_choices}

Generate a video ending on a white background with THREE lines of large, clearly readable, centered text:
  Line 1: the correct option letter (A/B/C/D).
  Line 2: the correct formula exactly as written above.
  Line 3: the formula with annotated values substituted in.

When substituting: if multiple objects exist, pick one consistent set.
\end{lstlisting}

\paragraph{Formulation-Graphic.}\leavevmode
As in the unified-model variant, \texttt{\{target\_time\}} is the case-specific target instant.
\begin{lstlisting}[style=prompt]
Predict the physical state at t = {target_time} s. Generate a video where the original annotations (text labels, coordinate axes, arrows) gradually fade out and cleanly disappear, while the background, ground, sky, and all scene lighting remain completely unchanged throughout. Objects move smoothly to their correct positions at t = {target_time} s -- each object must retain its exact texture, material, size, and shape. For each moving object, draw a borderless orange arrow starting from the object's center pointing in its velocity direction, with a label "v = X.XX" (speed value, no unit). If at rest, show "v = 0" with no arrow. Static camera, no zoom, no rotation.
\end{lstlisting}

\paragraph{Deduction.}\leavevmode
The runtime placeholder \texttt{\{physics\_duration\}} is the case's physics duration in seconds.
\begin{lstlisting}[style=prompt]
The first frame is an annotated physics scene. All annotations follow SI units: g is gravitational acceleration, e is restitution (0-1), v0 is initial velocity (arrows show direction), mu is friction coefficient. Generate {physics_duration} seconds of physically accurate motion strictly following these parameters. Every object AND the ground surface are perfectly rigid -- nothing breaks, cracks, deforms, or creates craters. Each object maintains its exact material, color, and texture throughout. All annotations must cleanly disappear after the first frame. Static camera, fixed background. Smooth, continuous motion.
\end{lstlisting}

\subsection{Protocol Ablations}
\label{app:protocol_ablations}

We use two ablations to verify that the protocol does not artificially inject the solution. In the
Text-Parameter ablation, physical quantities are supplied in structured text rather than as infographic
overlays. The small average change indicates that infographic annotations are not a shortcut, but a
spatially grounded way to bind quantities to visual referents. In the Concise-Prompt ablation, we
remove detailed output-format instructions. The main degradation appears on graphic-output tracks,
showing that the full prompt mainly standardizes answer format rather than providing physics
solutions.

\section{Evaluation Standardization Across Model Outputs}
\label{app:standardization}

\subsection{Output Canonicalization}
\label{app:output_canonicalization}

Different models return different output formats, resolutions, fps values, and durations. Before
scoring, every response is converted into a canonical evaluation packet, as shown in
Table~\ref{tab:canonicalization}.

\begin{table}[t]
\centering
\small
\caption{Canonical evaluation packets for different model families.}
\label{tab:canonicalization}
\begin{tabular}{p{0.22\linewidth}p{0.32\linewidth}p{0.32\linewidth}}
\toprule
Subtrack & Unified models & Video-generation models \\
\midrule
P-T / P-G / F-T / F-G & Generated PNG, evaluated directly as the answer artifact. &
Generated MP4 decoded; last frame used as the answer artifact. \\
Deduction & Timestamped keyframes generated at $\mathcal T_c$. &
Generated MP4 decoded; frames sampled after time-rescaling to the requested physical duration. \\
\bottomrule
\end{tabular}
\end{table}

\subsection{Time and FPS Alignment}
\label{app:time_alignment}

Each case $c$ defines a requested physical duration $T_c$ and a native GT frame rate
$f^{\mathrm{GT}}_c$. The GT frame rate is source-dependent: 24 fps for simulator cases, 30 fps for
self-recorded real-world cases, and the native fps stored in metadata for Internet-sourced cases.

For a GT frame sequence with $N^{\mathrm{GT}}_c$ decoded frames, the GT frame nearest to physical
time $t\in[0,T_c]$ is
\begin{equation}
i_c(t)=\mathrm{clip}\!\left(\mathrm{round}(t f^{\mathrm{GT}}_c),\ 0,\ N^{\mathrm{GT}}_c-1\right).
\end{equation}

For a generated video, we do not interpret the provider's raw output duration as physical time. The
prompt explicitly requests $T_c$ seconds of motion, so the decoded generated sequence is treated as a
continuous trajectory over $[0,T_c]$, regardless of whether the provider returns a nominal 5-second,
8-second, or other-length video. If the generated video contains $N^{\mathrm{gen}}$ frames, its
effective evaluation fps is
\begin{equation}
\widehat f^{\mathrm{gen}}=\frac{N^{\mathrm{gen}}}{T_c}.
\end{equation}
The generated frame nearest to physical time $t$ is then
\begin{equation}
j_c(t)=\mathrm{clip}\!\left(\mathrm{round}(t\widehat f^{\mathrm{gen}}),\ 0,\ N^{\mathrm{gen}}-1\right).
\end{equation}

This rule makes fps and duration comparable across models. Raw generated fps is used only for
decoding; physical time is defined by the prompt and the case metadata.

\paragraph{Example.}
Suppose the GT clip is 4 seconds at 30 fps, so it has 120 frames. A model returns an 8-second video
at 24 fps, so it has 192 decoded frames. Because the prompt requested ``Generate 4 seconds of
motion,'' we interpret the 192 generated frames as a 4-second trajectory and set
$\widehat f^{\mathrm{gen}}=192/4=48$ fps. At $t=2$ seconds, we compare GT frame
$i(2)=60$ with generated frame $j(2)=96$. The model's raw 8-second container duration is ignored
for physical-time alignment.

\subsection{Timestamp Grid}
\label{app:timestamp_grid}

Each case stores a canonical timestamp set $\mathcal T_c\subset[0,T_c]$. Deduction metrics compare
generated and GT frames at the same physical timestamps. For unified models, the timestamps in
$\mathcal T_c$ are the keyframes requested from the model. For video-generation models, the video is
decoded densely and sampled at the nearest generated frame $j_c(t)$ for each $t\in\mathcal T_c$.
When event-critical moments exist, such as collision contact or ramp exit, they are included in
$\mathcal T_c$ in addition to the uniform sampling grid.

\subsection{Resolution Normalization}
\label{app:resolution}

All pixel-level and mask-level metrics are computed at the GT resolution. If a generated image or
frame has size $H_{\mathrm{gen}}\times W_{\mathrm{gen}}$ and the GT frame has size
$H_{\mathrm{GT}}\times W_{\mathrm{GT}}$, the generated output is directly resized to
$H_{\mathrm{GT}}\times W_{\mathrm{GT}}$ before scoring. We use bilinear interpolation for RGB
frames and nearest-neighbor interpolation for masks. We do not crop, pad, or preserve aspect ratio
during metric computation; the same direct-resize rule is applied to every model to avoid
model-specific canvas handling.

\subsection{Graphic-Subtrack Mask IoU}
\label{app:graphic_iou}

Perception-Graphic and Formulation-Graphic have objective mask metrics. For P-G, the GT is the
object-only white-background artifact. For F-G, the GT is the target-time scene at $t^\star$. In both
cases, generated and GT masks are compared after resolution normalization.

For each matched object $o$, the mask IoU is
\begin{equation}
\mathrm{IoU}_o=\frac{|M^{\mathrm{gen}}_o\cap M^{\mathrm{GT}}_o|}
{|M^{\mathrm{gen}}_o\cup M^{\mathrm{GT}}_o|+\epsilon}.
\end{equation}
For multi-object cases, masks are matched by object identity when available and otherwise by
Hungarian matching over center distance and bounding-box overlap. The case-level IoU is the mean
over matched objects, with missing objects receiving zero IoU.

\subsection{Invalid Outputs and Rollout Aggregation}
\label{app:invalid_rollouts}

Each model is evaluated with three independent rollouts per case and subtrack. Let
$s_{m,c,k,r}$ be the score of model $m$ on case $c$, subtrack $k$, and rollout $r$. The rollout-mean
score is
\begin{equation}
S_{m,c,k}=\frac{1}{3}\sum_{r=1}^{3}s_{m,c,k,r}.
\end{equation}
If a rollout produces no decodable artifact, the rollout receives score zero for the affected subtrack.
Transient judge or parser failures are retried; persistent failures are recorded in the evaluation log
and assigned zero rather than removed from the denominator.

\section{Metric Definitions and Score Aggregation}
\label{app:metrics}

\subsection{MLLM-Judge Rubrics}
\label{app:mllm_rubric}

All MLLM judge calls use temperature 0. The judge sees the model output, the relevant reference
images or metadata, and a subtrack-specific rubric. Criteria are scored on $[0,1]$. Case-irrelevant
criteria are marked as not applicable and excluded from the group mean, with group weights
renormalized during score aggregation. Table~\ref{tab:rubrics} summarizes the rubric dimensions
used by each subtrack. The exact judge prompts, criterion-level fields, and group aggregation rules
are provided in Section~\ref{app:mllm_judge_prompts}.

\begin{table}[t]
\centering
\small
\caption{MLLM-judge rubric dimensions. P-G and F-G additionally include objective IoU as described
in Section~\ref{app:graphic_iou}.}
\label{tab:rubrics}
\begin{tabular}{p{0.16\linewidth}p{0.78\linewidth}}
\toprule
Subtrack & Rubric dimensions \\
\midrule
P-T & Annotation content, layout, and style/readability \\
P-G & Background purity, annotation removal, object preservation, and spatial consistency \\
F-T & Correct option, symbolic formula, numeric substitution, and presentation \\
F-G & Annotation removal, object match, velocity-arrow quality, and velocity text \\
Deduction & Annotation removal, object consistency, visual quality, motion smoothness, and physics accuracy \\
\bottomrule
\end{tabular}
\end{table}

For P-G, the spatial-consistency dimension incorporates the programmatic mask IoU from
Section~\ref{app:graphic_iou}. For F-G, the object-match dimension combines the MLLM visual
appearance score and the target-state mask IoU:
\begin{equation}
S_{\mathrm{obj}}^{\mathrm{F\text{-}G}}=
\frac{1}{2}S_{\mathrm{appearance}}^{\mathrm{MLLM}}+
\frac{1}{2}\mathrm{IoU}_{\mathrm{F\text{-}G}}.
\end{equation}
This keeps velocity-arrow and velocity-text requirements in the MLLM rubric while grounding object
position in an objective mask comparison.

\subsubsection{MLLM-Judge Prompt Templates}
\label{app:mllm_judge_prompts}

This section provides the MLLM-judge prompt templates used for subjective scoring.
For the four non-Deduction subtracks, unified understanding-generation models and
video-generation models are evaluated with the same MLLM-judge prompt templates.
The only difference lies in output canonicalization: unified models directly output the
final-frame image artifact, whereas video-generation models output an MP4 and the last
frame is extracted as the answer artifact.

For Deduction, the two model families use family-specific judge prompts because their
outputs have different temporal formats. Video-generation models are judged as full
videos using a video-level Deduction prompt. Unified models instead generate timestamped
keyframes and are judged frame-by-frame against the corresponding ground-truth frames
using a keyframe-level prompt. The keyframe-level prompt shares annotation removal,
object consistency, visual quality, and physics accuracy with the video-level prompt, but
uses object-position match against the timestamp-aligned ground-truth frame as the
keyframe proxy for temporal motion smoothness.

For each evaluated output, the judge is instructed to return only a JSON object. Each
scalar field is parsed as a score in the range [0,1], where 0 denotes a completely
incorrect or unusable response and 1 denotes a perfect match. A score of -1.0 is reserved
for criteria that are not applicable, such as velocity-arrow criteria when no velocity
arrow exists in the ground truth. In aggregation, inapplicable criteria are excluded from
the corresponding group mean, and the remaining group weights are renormalized. If a
response is undecodable or the judge output cannot be parsed after retries, the affected
rollout is assigned zero for the corresponding subtrack.

\paragraph{Perception-Text judge prompt.}
For Perception-Text, the judge compares the generated final-frame annotation artifact
against the ground-truth white-background annotation image.

{\footnotesize
\begin{verbatim}
You are an expert evaluator assessing how perfectly a generated image preserves the
exact annotation details from a ground truth reference image.

You are provided with two images:
1. "REFERENCE GROUND TRUTH IMAGE": The perfectly correct original image with
annotations on a pure white background.
2. "GENERATED IMAGE": The image to be evaluated.

Your task is to compare the GENERATED IMAGE against the REFERENCE GROUND TRUTH
IMAGE and score the generated image across 18 distinct criteria.
For each criterion, assign a score between 0.0 (completely unmatched/wrong) and 1.0
(perfectly matched).

Evaluate the following criteria carefully:

1. text_font_match: Does every text annotation use the EXACT SAME font family/typeface
as the ground truth?
2. text_style_match: Does every text annotation have the EXACT SAME style
(e.g., bold, italic, regular) as the ground truth?
3. text_size_match: Does every text annotation have the EXACT SAME font size/scale as
the ground truth?
4. text_color_match: Does every text annotation have the EXACT SAME color as the
ground truth?
5. text_position_match: Is every text annotation located in the EXACT SAME position
as the ground truth?
6. text_content_match: Does the textual content of every single annotation EXACTLY
match the ground truth (no typos, missing words, or hallucinations)?
7. text_readability: Are all text labels and annotations perfectly crisp, sharp, and
easy to read, without any blurriness, distortion, or generation artifacts, exactly
matching the clarity of the ground truth? (0.0=heavily blurred/distorted,
1.0=perfectly crisp and readable).
8. axes_color_match: Is the color of the coordinate axes arrows EXACTLY the same as
the ground truth?
9. axes_style_match: Is the line style (solid vs dashed) of the coordinate axes arrows
EXACTLY the same as the ground truth?
10. axes_size_match: Are the width, length, and arrowhead size of the coordinate axes
arrows EXACTLY the same as the ground truth?
11. axes_position_match: Is the position of the coordinate axes EXACTLY the same as
the ground truth?
12. axes_angle_match: Are the angles between the coordinate axes EXACTLY the same as
the ground truth?
13. velocity_color_match: If initial velocity arrows exist in the GT, does the generated
image have velocity arrows with the EXACT SAME color? (If no velocity arrows exist in
GT, score -1.0)
14. velocity_style_match: If initial velocity arrows exist in the GT, do they have the
EXACT SAME line style (solid vs dashed)? (If no velocity arrows exist in GT, score -1.0)
15. velocity_size_match: If initial velocity arrows exist in the GT, do they have the
EXACT SAME width, length, and arrowhead size? (If no velocity arrows exist in GT,
score -1.0)
16. velocity_position_match: If initial velocity arrows exist in the GT, are they in the
EXACT SAME starting position? (If no velocity arrows exist in GT, score -1.0)
17. no_extra_annotations: Does the generated image only contain the annotations present
in the GT, with absolutely NO extra, hallucinated, or duplicated annotations?
(0.0=extra stuff exists, 1.0=clean, matches GT exactly).
18. white_background: Is the background of the entire generated image pure white,
just like the ground truth?

Respond with ONLY JSON:
{
  "text_font_match": <float>,
  "text_style_match": <float>,
  "text_size_match": <float>,
  "text_color_match": <float>,
  "text_position_match": <float>,
  "text_content_match": <float>,
  "text_readability": <float>,
  "axes_color_match": <float>,
  "axes_style_match": <float>,
  "axes_size_match": <float>,
  "axes_position_match": <float>,
  "axes_angle_match": <float>,
  "velocity_color_match": <float>,
  "velocity_style_match": <float>,
  "velocity_size_match": <float>,
  "velocity_position_match": <float>,
  "no_extra_annotations": <float>,
  "white_background": <float>,
  "feedback": "<2-4 sentences summarizing any deviations found across these criteria>"
}
\end{verbatim}
}

The Perception-Text score is computed from three groups: content, layout, and style.
The content group averages text content match, text readability, and no-extra-annotation
score. The layout group averages text position, axes position, axes angle, and velocity
position. The style group averages text font, text style, text size, text color, axes
color, axes style, axes size, velocity color, velocity style, velocity size, and white
background. The final Perception-Text score is a weighted average of these three groups
with weights 0.50, 0.30, and 0.20.

\paragraph{Perception-Graphic judge prompt.}
For Perception-Graphic, the judge receives the annotation-only reference, the object-only
reference, and the generated object-only artifact.

{\footnotesize
\begin{verbatim}
You are an expert image evaluator.

You are provided with THREE images:
1. "REFERENCE WHITE BACKGROUND ANNOTATION IMAGE": a ground-truth reference on a
pure white background that contains human annotations (coordinate axes, text labels,
and possibly initial-velocity arrows).
2. "REFERENCE WHITE BACKGROUND OBJECT IMAGE": a ground-truth reference on a pure
white background that contains ONLY the physical objects (no annotations, no
background/ground/environment).
3. "GENERATED IMAGE": the image to be evaluated.

Your job is to score how well the GENERATED IMAGE matches the intended "objects-only
on pure white" result, using the two reference images to understand what should be
removed and what should remain.

Score each metric from 0.0 to 1.0 (higher is better). Be strict: partial visibility,
faint remnants, blur, or mismatched appearance should reduce the score.

BACKGROUND (purity):
- white_background: Is the entire background in the GENERATED IMAGE truly pure white
(no gradients, shadows, textures, ground plane, or environmental remnants)?

ANNOTATION REMOVAL (compare against the annotation reference):
- coord_axes: Are coordinate axes and axis labels fully removed?
(0=clearly visible, 1=completely gone)
- velocity_arrows: Are any initial-velocity arrows fully removed?
(0=visible, 1=gone. If NO velocity arrows exist in the annotation reference,
return -1.0.)
- text_annotations: Are ALL text labels, numbers, and other textual annotations removed?
(0=visible, 1=gone)

OBJECT PRESERVATION (compare against the object reference):
- objects_only_from_gt: Does the GENERATED IMAGE contain ONLY the physical objects
present in the object reference (no extra objects, no missing objects, and no leftover
ground/background surfaces)?
- object_completeness: Are all objects from the object reference present and
uncut/undamaged in the GENERATED IMAGE (no missing parts, no erasures, no merging into
the background)?
- object_visual_match: Do the kept objects match the object reference visually
(shape/geometry, relative size, colors, textures/materials, shading/lighting cues, and
any distinctive visual details)? (0=looks different, 1=visually identical)
- object_position_match: Do the objects appear at the EXACT SAME image positions as in
the object reference (same 2D placement and relative arrangement; no shifting, drifting,
or rearrangement)?

Respond with ONLY JSON:
{
  "white_background": <float>,
  "coord_axes": <float>,
  "velocity_arrows": <float>,
  "text_annotations": <float>,
  "objects_only_from_gt": <float>,
  "object_completeness": <float>,
  "object_visual_match": <float>,
  "object_position_match": <float>,
  "feedback": "<2-4 sentences summarizing the biggest issues>"
}
\end{verbatim}
}

The Perception-Graphic score is computed from four groups: background purity, annotation
removal, object preservation, and spatial consistency. The background group contains
white background. The annotation-removal group averages coordinate axes, velocity arrows,
and text annotations. The object-preservation group averages objects-only-from-ground
truth, object completeness, and object visual match. The spatial group averages object
position match and the programmatic segmentation IoU. The final Perception-Graphic score
uses group weights 0.10, 0.30, 0.30, and 0.30.

\paragraph{Formulation-Text judge prompt.}
For Formulation-Text, the judge evaluates whether the final white-background answer
contains the correct option, the correct symbolic law, and a correct numeric substitution.

{\footnotesize
\begin{verbatim}
You are evaluating a model's physical comprehension.

The model was asked to output an image containing 3 text lines on a pure white background:
Line 1: The correct option letter.
Line 2: The correct formula.
Line 3: The formula with annotated variables replaced by corresponding values. If the
scene has multiple objects, values belonging to specific objects must come from exactly
ONE consistent set (e.g., one object for single-body motion, or multiple objects forming
one valid set for multiple-body motion).

Correct Formula is {correct_letter}: {correct_formula}
Annotation Content: {annotation}

Evaluate the generated image (Score 0.0-1.0):

- option_correct: Is the correct letter (Line 1) chosen?
- formula_variables_correct: Are physical symbols in the formula (Line 2) written
correctly?
- formula_constants_correct: Are inherent numeric constants in the formula written
correctly?
- formula_operators_correct: Are operators in the formula (Line 2) written correctly?
- substitution_all_replaced: Are ALL annotated variables replaced by values in Line 3?
- substitution_values_correct: Are substituted values correct? If there are multiple
objects, did the model pick exactly one valid set of objects to substitute values?
- substitution_unreplaced_correct: Are unannotated symbols correctly kept unchanged in
Line 3?
- substitution_constants_correct: Are inherent numeric constants correctly kept in
Line 3?
- substitution_operators_correct: Are operators correctly written in Line 3?
- format_3_lines: Does it strictly follow the 3-line format?
- background_pure_white: Is the background pure white?

Respond with ONLY JSON:
{
  "option_correct": <float>,
  "formula_variables_correct": <float>,
  "formula_constants_correct": <float>,
  "formula_operators_correct": <float>,
  "substitution_all_replaced": <float>,
  "substitution_values_correct": <float>,
  "substitution_unreplaced_correct": <float>,
  "substitution_constants_correct": <float>,
  "substitution_operators_correct": <float>,
  "format_3_lines": <float>,
  "background_pure_white": <float>,
  "feedback": "<2-3 sentences summarizing issues>"
}
\end{verbatim}
}

The Formulation-Text score is computed from four groups: correct option, symbolic
formula, numeric substitution, and presentation. The symbolic-formula group averages
formula variables, constants, and operators. The numeric-substitution group averages
whether all annotated variables are replaced, whether substituted values are correct,
whether unannotated symbols are preserved, whether constants are preserved, and whether
operators are correct. The presentation group averages the three-line format score and
the pure-white-background score. The final Formulation-Text score uses group weights
0.20, 0.30, 0.40, and 0.10.

\paragraph{Formulation-Graphic judge prompt.}
For Formulation-Graphic, the judge compares the generated target-time scene against
the ground-truth target-time reference and the initial annotated scene.

{\footnotesize
\begin{verbatim}
You are evaluating a generated image against a ground truth (GT) reference image and an
initial annotated scene image.

The images are provided in this order:
1. REFERENCE GROUND TRUTH IMAGE (showing the correct physical positions of all objects,
their velocity arrow directions, and their instantaneous speed values).
2. INITIAL ANNOTATED PHYSICS SCENE (showing objects, axes, and all initial text labels).
3. GENERATED IMAGE.

Please evaluate the GENERATED IMAGE based on the following metrics. Score each from
0.0 to 1.0. If there are multiple objects, your score should reflect the average
performance across ALL objects:

- annotations_removed: Are all annotations from the INITIAL ANNOTATED PHYSICS SCENE
(coordinate axes, text labels, initial velocity arrows) correctly removed in the
GENERATED IMAGE? (0=not removed at all, 1=completely clean)
- object_position_match: Are the positions of all objects in the GENERATED IMAGE exactly
the same as their corresponding positions in the GT reference image?
(0=completely different, 1=exact match for all objects)
- object_appearance_match: Do all objects in the GENERATED IMAGE look visually identical
to the corresponding objects in the GT reference image, with special attention to EXACT
texture/material detail and EXACT object size/scale? (0=very different, 1=identical for
all objects)
- arrow_is_borderless_orange: Are the instantaneous velocity arrows for all moving
objects in the GENERATED IMAGE borderless orange arrows? (0=no/wrong style/color for all,
1=yes for all)
- arrow_from_center: Do the instantaneous velocity arrows in the GENERATED IMAGE
originate from the centers of their respective objects? (0=no for all, 1=yes for all)
- arrow_direction_match: Are the directions of the instantaneous velocity arrows in the
GENERATED IMAGE completely consistent with the corresponding velocity arrow directions
in the GT reference image? (0=completely wrong directions, 1=exact match for all moving
objects)
- velocity_label_present_and_format: Does EVERY object in the GENERATED IMAGE have a
velocity text label that roughly follows the pattern "v = number" (or "v = 0" if at rest)?
The exact spacing/decimals do NOT need to match; the key is that a clear velocity label
exists. (0=missing labels for all objects, 1=labels present for all objects)
- velocity_value_match: How closely do the numerical values of the velocity text labels
in the GENERATED IMAGE match the values in the GT reference image? (0=values are
completely wrong or missing, 1=values match exactly for all objects. Give partial credit,
e.g., 0.1-0.9, depending on how numerically close the values are and how many objects are
correct)
- velocity_label_no_unit: Do the velocity text labels in the GENERATED IMAGE contain only
numbers WITHOUT any units (e.g., no "m/s" or similar)? (0=all have units, 1=no units for
any object)

Respond with ONLY JSON:
{
  "annotations_removed": <float>,
  "object_position_match": <float>,
  "object_appearance_match": <float>,
  "arrow_is_borderless_orange": <float>,
  "arrow_from_center": <float>,
  "arrow_direction_match": <float>,
  "velocity_label_present_and_format": <float>,
  "velocity_value_match": <float>,
  "velocity_label_no_unit": <float>,
  "feedback": "<2-4 sentences summarizing issues>"
}
\end{verbatim}
}

The Formulation-Graphic score is computed from four groups: annotation removal, object
match, velocity-arrow quality, and velocity text. The object-match group averages object
position match, object appearance match, and the target-state segmentation IoU. The
velocity-arrow group averages arrow style/color, arrow origin, and arrow direction. The
velocity-text group averages velocity-label presence and format, velocity-value match,
and absence of units. The final Formulation-Graphic score uses group weights 0.10, 0.30,
0.30, and 0.30.

\paragraph{Deduction judge prompt for video-generation models.}
For video-generation models, the judge evaluates the full generated video. When
ground-truth frames are available to the judge, the optional ground-truth comparison
section is inserted through the placeholder.

{\footnotesize
\begin{verbatim}
You are evaluating a physics video generation model's ability to generate physically
accurate and visually high-quality motion.

You are provided with:
1. INITIAL ANNOTATED PHYSICS SCENE (the first frame, showing objects, coordinate axes,
and annotated initial conditions).
2. GENERATED VIDEO (the model's output).
{gt_section}

Please evaluate the GENERATED VIDEO based on the following metrics. Score each from
0.0 to 1.0:

- gen_annotations_removed: Are all human annotations from the INITIAL SCENE (such as
text labels, coordinate axes, and initial velocity arrows) successfully removed in the
generated video? (0 = annotations remain visible or leave severe ghostly artifacts,
1 = perfectly clean removal, leaving only the physical objects and background).
- object_consistency: Do the physical objects remain perfectly consistent throughout the
video? (0 = objects suddenly disappear, new objects appear out of nowhere, or objects
undergo severe unintended deformation, 1 = objects maintain their exact identity, shape,
and count perfectly).
- visual_quality: What is the overall visual quality of the generated video? Consider
sharpness, clarity, and absence of visual artifacts or noise. (0 = severe artifacts or
blur, 1 = high-quality, crisp rendering).
- motion_smoothness: Is the temporal motion of the objects smooth and natural?
(0 = severely broken motion, flickering, stuttering, or teleporting, 1 = perfectly
smooth, continuous, and natural movement).
- physics_accuracy: Does the motion of the objects accurately follow the physics
described by the initial conditions (trajectory, velocity, gravity, collisions, etc.)?
(0 = completely defies physics, 1 = physically accurate motion).

Respond with ONLY JSON:
{
  "gen_annotations_removed": <float>,
  "object_consistency": <float>,
  "visual_quality": <float>,
  "motion_smoothness": <float>,
  "physics_accuracy": <float>,
  "feedback": "<2-4 sentences summarizing issues regarding annotation removal, visual
  quality, object consistency, smoothness, and physics.>"
}
\end{verbatim}
}

The optional ground-truth comparison section is:

{\footnotesize
\begin{verbatim}
3. GROUND TRUTH FRAMES showing the correct physics are also provided. Compare generated
vs GT: object positions, trajectories, timing.
\end{verbatim}
}

\paragraph{Deduction keyframe judge prompt for unified understanding-generation models.}
For unified understanding-generation models that output timestamped keyframes rather
than dense videos, the judge evaluates each generated keyframe against the corresponding
ground-truth frame.

{\footnotesize
\begin{verbatim}
You are evaluating a single keyframe from a physics simulation generation model.

You are provided with:
1. INITIAL ANNOTATED PHYSICS SCENE (the first frame with coordinate axes, text labels,
etc.)
2. GROUND TRUTH FRAME at t = {time_point:.2f}s (the correct physical state at this
moment)
3. GENERATED FRAME at t = {time_point:.2f}s (the model's output for this moment)

Compare the GENERATED FRAME against the GROUND TRUTH FRAME. Score each from 0.0 to 1.0:

- gen_annotations_removed: Are all human annotations from the INITIAL SCENE (text labels,
coordinate axes, initial velocity arrows) removed in the generated frame?
(0 = annotations clearly visible, 1 = perfectly clean)
- object_consistency: Do the physical objects match the GT in identity, count, and shape?
(0 = objects missing/hallucinated/severely deformed, 1 = exact match)
- visual_quality: Overall visual quality -- sharpness, clarity, absence of artifacts.
(0 = severe artifacts/blur, 1 = high quality crisp rendering)
- object_position_match: Are the objects at the same positions as in the GT frame?
(0 = completely wrong positions, 1 = exact position match)
- physics_accuracy: Does the overall physical state (positions, deformations,
interactions) match the GT? (0 = completely wrong, 1 = physically accurate)

Respond with ONLY JSON:
{
  "gen_annotations_removed": <float>,
  "object_consistency": <float>,
  "visual_quality": <float>,
  "object_position_match": <float>,
  "physics_accuracy": <float>,
  "feedback": "<1-2 sentences on the biggest issues>"
}
\end{verbatim}
}

For Deduction, the subjective MLLM scores are combined with programmatic objective
metrics. For video-generation models, annotation removal and object consistency form the
integrity group; visual quality, motion smoothness, normalized PSNR, and masked PSNR form
the fidelity group; physics accuracy, Spatial IoU, Spatiotemporal IoU, Weighted Spatial
IoU, and velocity accuracy form the physics group. The final Deduction score uses group
weights 0.20 for integrity, 0.20 for fidelity, and 0.60 for physics.

For unified understanding-generation models, the keyframe-level scores are first averaged
across the requested timestamps. The object-position-match field is then used as the
keyframe-level proxy for the shared motion-smoothness field before applying the same
Deduction score aggregation.

\subsection{Deduction Objective Metrics}
\label{app:deduction_metrics}

Deduction is evaluated at the timestamp set $\mathcal T_c$. For every timestamp, the generated frame
is selected by the time-rescaling rule in Section~\ref{app:time_alignment}, resized to GT resolution,
and compared against the corresponding GT frame.

\paragraph{Normalized PSNR.}
We compute frame-level PSNR between generated and GT RGB frames and normalize by clipping at
40 dB:
\begin{equation}
S_{\mathrm{PSNR}}=\min(\mathrm{PSNR}/40,1).
\end{equation}

\paragraph{Masked PSNR.}
Masked PSNR is computed only on the union of GT foreground object masks. This reduces background
dominance and focuses the metric on object placement and appearance.

\paragraph{Motion-mask extraction.}
Following Physics-IQ~\cite{Motamed_2026_WACV}, the three IoU-based
Deduction metrics are computed on \emph{motion/action masks}, not on raw RGB
frames. After applying the time-rescaling rule of Section~\ref{app:time_alignment}
and resizing generated frames to the GT resolution, we use SAM3 to segment the
physically relevant moving objects in both the generated clip and the GT clip,
yielding binary mask sequences
\[
\mathcal A^{\mathrm{gen}}=\{A^{\mathrm{gen}}_t\}_{t\in\mathcal T_c},
\qquad
\mathcal A^{\mathrm{GT}}=\{A^{\mathrm{GT}}_t\}_{t\in\mathcal T_c}.
\]
Here $A_t(p)\in\{0,1\}$ indicates whether pixel $p$ belongs to a
physically relevant moving object at physical time $t$. For simulator cases,
the engine-provided instance masks are used as GT masks when available; for
real-world cases, the human-refined SAM3 masks are used as GT masks. For
generated videos, SAM3 is applied with the same object prompts and
post-processing rules as the GT annotation pipeline. The resulting masks are
therefore object/action masks: they measure the spatial support of the
law-relevant moving bodies, rather than exact RGB texture or background
appearance.

\paragraph{Spatial IoU.}
Spatial IoU answers the Physics-IQ question: \emph{where does action happen?}
For each clip, we collapse the binary motion-mask sequence along the time
dimension with a max operation:
\[
S^{x}(p)=\max_{t\in\mathcal T_c} A^{x}_t(p),
\qquad x\in\{\mathrm{gen},\mathrm{GT}\}.
\]
Thus $S^{x}(p)=1$ if any motion occurs at pixel $p$ at any evaluated time,
and $0$ otherwise. The Spatial IoU is then
\[
S_{\mathrm{spatialIoU}}
=
\frac{
\sum_{p} \mathbf{1}\!\left[S^{\mathrm{gen}}(p)=1
\wedge S^{\mathrm{GT}}(p)=1\right]
}{
\sum_{p} \mathbf{1}\!\left[S^{\mathrm{gen}}(p)=1
\vee S^{\mathrm{GT}}(p)=1\right]
+\epsilon
}.
\]
This metric ignores the exact timing of the action and evaluates whether the
generated video places motion in the same spatial regions as the GT video.

\paragraph{Spatiotemporal IoU.}
Spatiotemporal IoU answers the Physics-IQ question:
\emph{where and when does action happen?} Instead of collapsing time, we
compare the generated and GT motion masks at each aligned physical timestamp
and average the resulting frame-wise IoUs:
\[
S_{\mathrm{stIoU}}
=
\frac{1}{|\mathcal T_c|}
\sum_{t\in\mathcal T_c}
\frac{
\sum_p \mathbf{1}\!\left[A^{\mathrm{gen}}_t(p)=1
\wedge A^{\mathrm{GT}}_t(p)=1\right]
}{
\sum_p \mathbf{1}\!\left[A^{\mathrm{gen}}_t(p)=1
\vee A^{\mathrm{GT}}_t(p)=1\right]
+\epsilon
}.
\]
A model can therefore obtain high Spatial IoU but low Spatiotemporal IoU if
it produces action in the right places but at the wrong time.

\paragraph{Weighted Spatial IoU.}
Weighted Spatial IoU answers the Physics-IQ question:
\emph{where and how much action happens?} Rather than using a binary
``ever-moved'' spatial map, we collapse each motion-mask sequence by temporal
averaging:
\[
W^{x}(p)=\frac{1}{|\mathcal T_c|}
\sum_{t\in\mathcal T_c} A^{x}_t(p),
\qquad x\in\{\mathrm{gen},\mathrm{GT}\}.
\]
Here $W^{x}(p)\in[0,1]$ measures how frequently pixel $p$ is active over the
evaluated clip. The weighted spatial overlap is computed by the generalized
IoU ratio used in Physics-IQ:
\[
S_{\mathrm{wIoU}}
=
\frac{
\sum_p \min\!\left(W^{\mathrm{gen}}(p), W^{\mathrm{GT}}(p)\right)
}{
\sum_p \max\!\left(W^{\mathrm{gen}}(p), W^{\mathrm{GT}}(p)\right)
+\epsilon
}.
\]
Unlike Spatial IoU, this metric distinguishes transient motion from repeated
or prolonged motion at the same location. For example, a ball passing through
a region once and an object repeatedly oscillating through that region may
share a similar binary spatial support, but they induce different weighted
motion maps.

\paragraph{Velocity accuracy.}
For sampled frames where 3D velocity can be estimated, we recover object-level velocity vectors and
compare them with the law-predicted velocity. If the mean Euclidean velocity error is $e_v$, the
velocity score is
\begin{equation}
S_{\mathrm{vel}}=\frac{1}{1+e_v}.
\end{equation}
This maps larger errors smoothly toward zero while preserving a $[0,1]$ score range.

\subsection{Deduction Score Fusion}
\label{app:deduction_score}

The final Deduction score combines MLLM and objective metrics in three groups:
\begin{align}
S_{\mathrm{integrity}} &=
\frac{1}{2}\left(S_{\mathrm{ann\_removed}}+S_{\mathrm{object\_consistency}}\right),\\
S_{\mathrm{fidelity}} &=
\frac{1}{4}\left(S_{\mathrm{visual\_quality}}+S_{\mathrm{motion\_smoothness}}
+S_{\mathrm{PSNR}}+S_{\mathrm{maskedPSNR}}\right),\\
S_{\mathrm{physics}} &=
\frac{1}{5}\left(S_{\mathrm{physics\_accuracy}}+S_{\mathrm{spatialIoU}}
+S_{\mathrm{stIoU}}+S_{\mathrm{wIoU}}+S_{\mathrm{vel}}\right).
\end{align}
The Deduction score is
\begin{equation}
S_{\mathrm{Ded}}=0.20S_{\mathrm{integrity}}+
0.20S_{\mathrm{fidelity}}+
0.60S_{\mathrm{physics}}.
\end{equation}
The larger physics weight reflects the benchmark objective: visual plausibility should not compensate
for physically incorrect dynamics.

\subsection{Hierarchical Aggregation}
\label{app:aggregation}

Scores are aggregated from rollout to case, from case to slice, and from slice to model. The three
rollouts are first averaged as in Section~\ref{app:invalid_rollouts}. For a model $m$ and subtrack
$k$, the track-wise score is
\begin{equation}
S_{m,k}=\frac{1}{|\mathcal C|}\sum_{c\in\mathcal C}S_{m,c,k}.
\end{equation}
The overall model score averages over cases and subtracks:
\begin{equation}
S_m=\frac{1}{|\mathcal C||\mathcal K|}\sum_{c\in\mathcal C}\sum_{k\in\mathcal K}S_{m,c,k}.
\end{equation}
Pillar-wise and source-wise scores are computed by restricting $\mathcal C$ to the corresponding
case subset. All reported cells are case-level means over the relevant subset, not means of previously
reported table cells.

\section{Judge Reliability and Robustness}
\label{app:judge_reliability}

\subsection{Primary Judge}
\label{app:primary_judge}

The canonical MLLM judge is Gemini 3 Flash with temperature 0. The judge is used for response
validity, format compliance, text rendering, visual quality, object consistency, velocity-arrow
interpretation, and high-level physics plausibility. Programmatic metrics are used wherever masks,
pixels, or velocities can be compared directly to GT.

\subsection{Open-Weights Cross-Check}
\label{app:judge_crosscheck}

To verify that the leaderboard is not an artifact of a single judge, we re-evaluate a balanced subset of
outputs using Qwen3-VL-30B-A3B-Instruct-FP8 as an open-weights judge. The cross-check covers
seven representative candidate generation models and all five subtracks. The goal is ranking
robustness rather than exact equality of absolute scores, since different judges may calibrate the
$[0,1]$ scale differently.

Table~\ref{tab:judge_agreement} reports the agreement. The two judges agree strongly on per-model
average scores and preserve the same quality-tier ordering.

\begin{table}[t]
\centering
\small
\caption{Agreement between Gemini 3 Flash and Qwen3-VL on per-model mean scores.}
\label{tab:judge_agreement}
\begin{tabular}{p{0.45\linewidth}p{0.18\linewidth}p{0.18\linewidth}}
\toprule
Comparison & Pearson $r$ & Spearman $\rho$ \\
\midrule
Overall model average & 0.95 & 0.93 \\
Formulation-Text & 0.99 & 1.00 \\
Deduction & 0.99 & 1.00 \\
Formulation-Graphic & 0.97 & 0.93 \\
Perception-Text & 0.81 & 0.96 \\
Perception-Graphic & 0.81 & 0.71 \\
\bottomrule
\end{tabular}
\end{table}

The residual disagreement is concentrated in two cases. First, the open-weights judge is sometimes
more lenient toward visually plausible but law-violating motion. Second, it can score rubric items
more independently, whereas the primary judge more consistently propagates earlier errors such as a
wrong formula option into downstream formula and substitution criteria. Because the tier-level ranking
is preserved, we treat Qwen3-VL as a reproducible robustness check and keep Gemini 3 Flash as the
canonical judge for the main leaderboard.

\section{Experimental Details}
\label{app:experiment_release}

The benchmark evaluates 11 models: five video-generation models and six unified
understanding-generation models. Each model is evaluated on 400 cases, five subtracks, and three
independent rollouts, for 6000 responses per model. We do not tune prompts per model. Models
receive the same annotated first frame and the same subtrack prompt template within their output
family.

For the four non-Deduction subtracks, unified models directly generate the answer artifact and video
models generate a chain-of-frames clip whose last frame is taken as the answer. For Deduction, video
models generate a full clip and unified models generate timestamped keyframes at the evaluation
timestamps.

\section{Limitations}
\label{app:limitations}

\paragraph{Physics scope.}
\applepi \ focuses on classical rigid-body mechanics: gravity-driven motion, momentum-conservation
collisions, Newton's first law, and their simple compositions. It does not evaluate fluids,
thermodynamics, electromagnetism, deformable bodies, fracture, granular media, biological motion,
or quantum phenomena.

\paragraph{Object and scene scope.}
Dynamic objects are restricted to sphere, cube, cylinder, and cone. This improves annotation
consistency but does not stress articulated, organic, deformable, or highly irregular objects. Cases use
a single fixed camera, so the benchmark does not directly test multi-view consistency or novel-view
generation.

\paragraph{Interface scope.}
The annotated first frame removes unnecessary reference ambiguity and makes physical quantities
spatially grounded. As a result, \applepi \ does not test whether a model can construct the full physical
scene from text alone. It tests whether, given a visually grounded physical setup, the model can
perceive quantities, formulate laws, and deduce dynamics.

\paragraph{Evaluation scope.}
MLLM judges are useful for response validity, format compliance, and perceptual qualities, but they
are not perfect physical oracles. We therefore combine them with objective metrics whenever masks,
pixels, or velocities can be compared to ground truth. Objective metrics also have limitations: mask
IoU depends on segmentation quality, pixel metrics can penalize harmless appearance differences,
and velocity estimation can be noisy for real-world clips.

\paragraph{Real-world ground-truth noise.}
Simulator cases have exact engine-state ground truth, but real-world cases rely on measurement,
mask refinement, and frame-based annotation. We mitigate this with controlled recording, instrument
precision, multi-pass review, and inter-annotator agreement analysis, but real-world labels remain
less exact than simulator labels.

\paragraph{Temporal normalization assumption.}
For generated videos, we interpret the full decoded output as the requested physical duration in the
prompt. This is necessary because providers may return fixed-length containers independent of the
requested duration. The rule is consistent across all models, but it assumes that the generated clip
represents the requested physical time rather than the provider's nominal file duration.

\end{document}